\documentclass[]{article}

\usepackage{microtype}
\usepackage{graphicx}
\usepackage{subfigure}
\usepackage{booktabs} 
\usepackage{natbib}
\usepackage{hyperref}

\usepackage{amsmath}
\usepackage{amssymb}
\usepackage{mathtools}
\usepackage{amsthm}

\usepackage[capitalize,noabbrev]{cleveref}

\theoremstyle{plain}

\theoremstyle{definition}

\theoremstyle{remark}

\usepackage[textsize=tiny]{todonotes}
\usepackage{interval}

\usepackage{thmtools, thm-restate}
\usepackage{rotating}
\usepackage{xcolor}
\usepackage{bbold}
\usepackage{tikz}
\usetikzlibrary{arrows}
\usetikzlibrary{calc}
\usetikzlibrary{fit}

\usepackage{rotating}

\newcommand{\bldr}{\mathbf{r}}

\newcommand{\bldz}{\mathbf{z}}

\newcommand{\blde}{\mathbf{e}}
\newcommand{\bldv}{\mathbf{v}}
\newcommand{\bldV}{\mathbf{V}}
\newcommand{\bldx}{\mathbf{x}}

\newcommand{\bldw}{\mathbf{w}}

\newcommand{\bldy}{\mathbf{y}}

\newcommand{\bldtheta}{\mathbf{\theta}}
\newcommand{\bldphi}{\mathbf{\phi}}
\newcommand{\bldzero}{\mathbf{0}}
\newcommand{\RR}{\mathbb{R}}
\newcommand{\NN}{\mathbb{N}}
\newcommand{\ZZ}{\mathbb{Z}}

\newcommand{\M}{\mathcal{M}}
\newcommand{\N}{\mathcal{N}}
\newcommand{\calS}{\mathcal{S}}

\newcommand{\smashsum}[1]{\smashoperator{\sum_{#1}}}
\def\ceil#1{\lceil #1 \rceil}
\def\floor#1{\lfloor #1 \rfloor}

\title{Global Optimization Networks}
\author{Sen Zhao\footnote{senzhao@google.com}, Erez Louidor\footnote{erez@google.com}, Olexander Mangylov\footnote{valaeda@gmail.com}, Maya Gupta\footnote{relativeentropy@gmail.com} \\ Google Research}

\begin{document}

\maketitle




\begin{abstract}
We consider the problem of estimating a good maximizer of a black-box function given noisy examples. To solve such problems, we propose to fit a new type of function which we call a global optimization network (GON), defined as any composition of an invertible function and a unimodal function, whose unique global maximizer can be inferred in $\mathcal{O}(D)$ time. In this paper, we show how to construct invertible and unimodal functions by using linear inequality constraints on lattice models. We also extend to \emph{conditional} GONs that find a global maximizer conditioned on specified inputs of other dimensions. Experiments show the GON maximizers are statistically significantly better predictions than those produced by convex fits, GPR, or DNNs, and are more reasonable predictions for real-world problems.
\end{abstract}

\section{Introduction}
We consider the problem of predicting a maximizer $\hat{\bldx}$ for an unknown function $g(\bldx): \mathbb{R}^D \rightarrow \mathbb{R}$, given only a fixed set of $N$ noisy input-output training pairs $(\bldx_i, y_i)$ for $\bldx_i \in \mathbb{R}^D$, and $y_i = g(\bldx_i) + \epsilon_i \in \RR, i=1, \dots, N$, where $\epsilon_i$ is zero-mean noise. The predicted maximizer $\hat{\bldx}$ will be judged by how close its predicted output $g(\hat{\bldx})$ is to the true global maximum $g(\bldx^*)$ where $\bldx^* \in \arg \max_\bldx g(\bldx)$. 

A few example applications of this \emph{global optimization} problem are predicting how many books to print of a new book run to maximize first year profit, predicting the optimal college for a particular student to maximize their happiness ten years later, predicting the optimal dosage of the medicine levothyroxine for an individual to minimize deviation from target TSH levels, and in general, predicting the optimal design specs a business should target when developing a new product to maximize sales.  


We leave as future work extending the proposed methodology to the standard \emph{global optimization algorithm} setting where one is allowed to make a series of guesses \citep{Pardalos}, that is, where one selects an $\hat{\bldx}_t$ and is able to acquire the additional training label $g(\hat{\bldx}_t) + \epsilon_t$, for $t = 1, \ldots, T$. Here, we make only one guess $\hat{x}$.

We take a machine-learning approach: we fit a function $h(\bldx;\bldphi)$ with  parameters $\bldphi$ to the $N$ training samples, and then take the maximizer of the fitted $h(\bldx;\bldphi)$ as $\hat{\bldx}$. A key question is which function class to use for $h(\bldx;\bldphi)$.  A suitable $h(\bldx;\bldphi)$ should have the right amount of model expressibility,  and ideally it will be easy to find its maximizer. \citet{BoxWilson:1951} proposed fitting a quadratic function as a surrogate function whose maximizer could then be easily found, but for many real-world applications a quadratic function will be too inflexible. At the other extreme, one can fit an arbitrarily flexible function like a DNN \citep{Gorissen:2010}, but that may overfit, producing a noisy estimate of the maximizer. In addition, it may be prohibitively expensive to find the predicted maximizer of a fitted flexible model like a DNN for even a small number of features $D$.

We propose a new function class which we call a \emph{global optimization network} (GON) that generalizes unimodal functions (for intuition, see the 1D example in Fig. \ref{fig:kings}). GONs aree more flexible than prior restricted surrogate functions, but have a well-defined global maximizer that can be found surprisingly efficiently in $\mathcal{O}(D)$ time for $D$ inputs.  We also extend GONs to the conditional setting where some of the inputs $\bldz\in\RR^M$ are fixed, and define \emph{conditional global optimization networks} (CGONs) that infer the conditional maximizer  $\bldx^* = \arg \max_\bldx g(\bldx, \bldz)$. 


GONs can be built using various choices of functions for each layer.  Specifically, we focus on showing how to construct GONs using constrained deep lattice networks (DLNs) \cite{You:2017}. A key benefit of a DLN GON is that its $D$-dimensional global maximizer can be found in $\mathcal{O}(D)$ time, and they can be trained efficiently by constrained empirical risk minimization with linear inequality constraints. 

\section{Related Work}\label{sec:relatedwork}
GONs lie at the intersection of two classic strategies: \emph{(i)} fitting models to noisy data to predict a maximizer of an unknown function, and \emph{(ii)} defining a function class by its \emph{shape constraints}. We survey those two strategies in the Appendix, and detail the closest related work next. 

\subsection{Closest Related Work on Function Fitting}
\citet{Zico:2017} proposed fitting a \emph{convex} deep neural network as a surrogate functions to predict a global minimizer $x^* = \arg \min_\bldx g(\bldx)$, which they constructed as a multi-layer ReLU net with the necessary monotonicity shape constraints to produce an overall convex function; this is called ICNN and sometimes FICNN. However, we found that convex functions were often too inflexible for this task, see Fig. \ref{fig:kings} for a 1D example. 

\citet{Zico:2017} also proposed \emph{partial-input convex neural network} (PICNN) for the \emph{conditional} global optimization problem $\bldx^* = \arg \min_\bldx g(\bldx, \bldz)$. Because both FICNN and PICNN is convex in $\bldx$, the fits can be minimized numerically to find $\arg\min_\bldx h(\bldx;\bldphi)$ and $\arg\min_\bldx h(\bldx,\bldz;\bldphi)$.  However, ReLU-activated ICNNs are neither smooth nor strongly convex, which reduces the convergence rate in finding the minimizer. 



\subsection{Closest Related Work in Shape Constraints}
\begin{figure}
\centering
\begin{tabular}{cccc}
\includegraphics[width=1in]{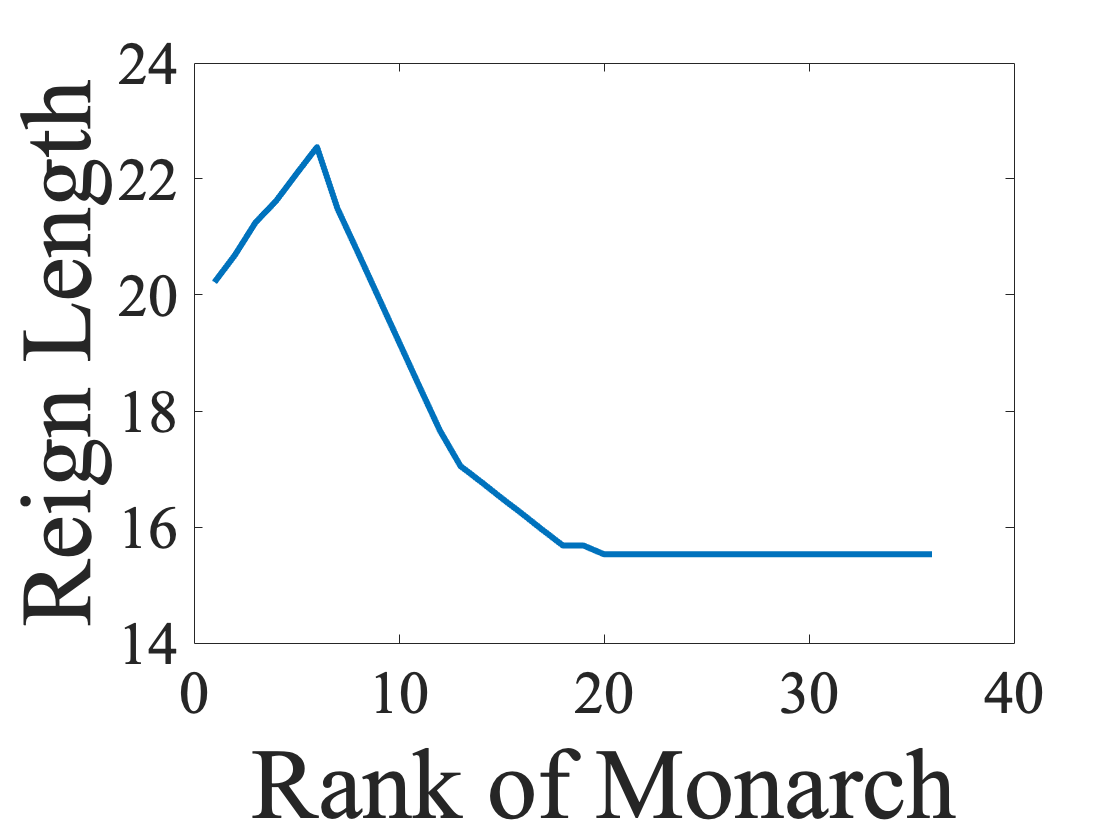}  &
\includegraphics[width=1in]{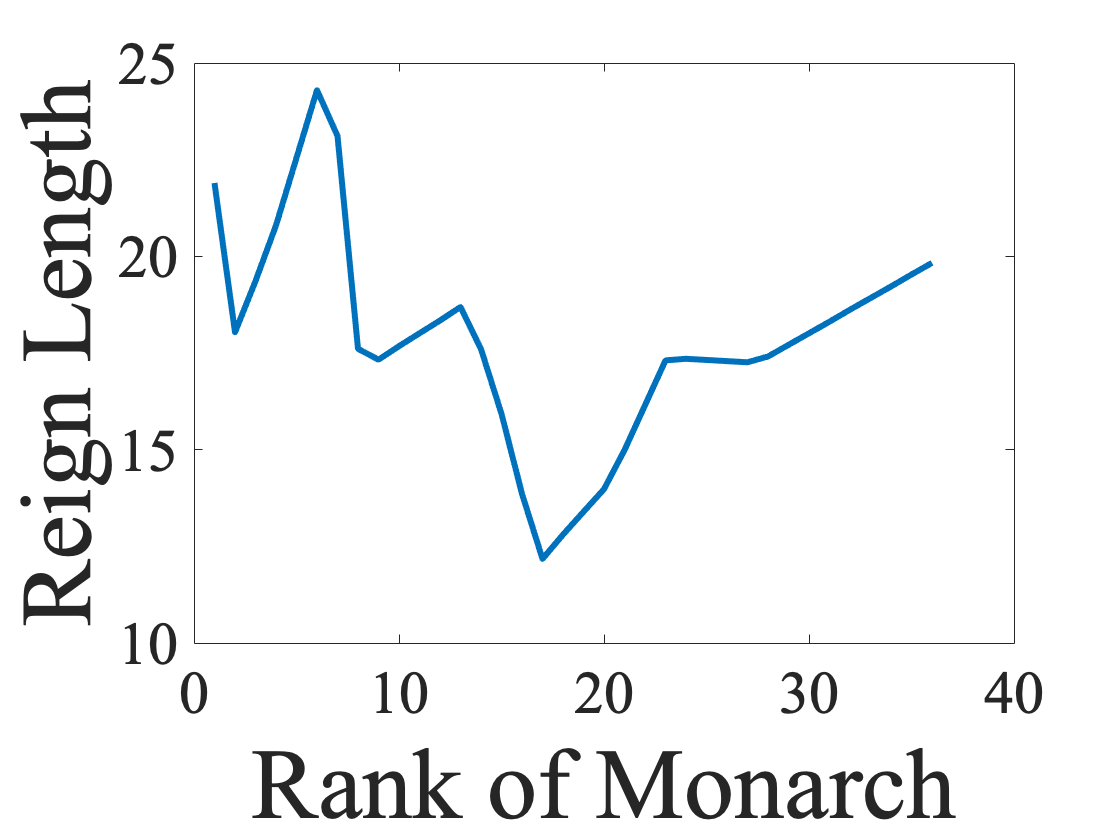} &
\includegraphics[width=1in]{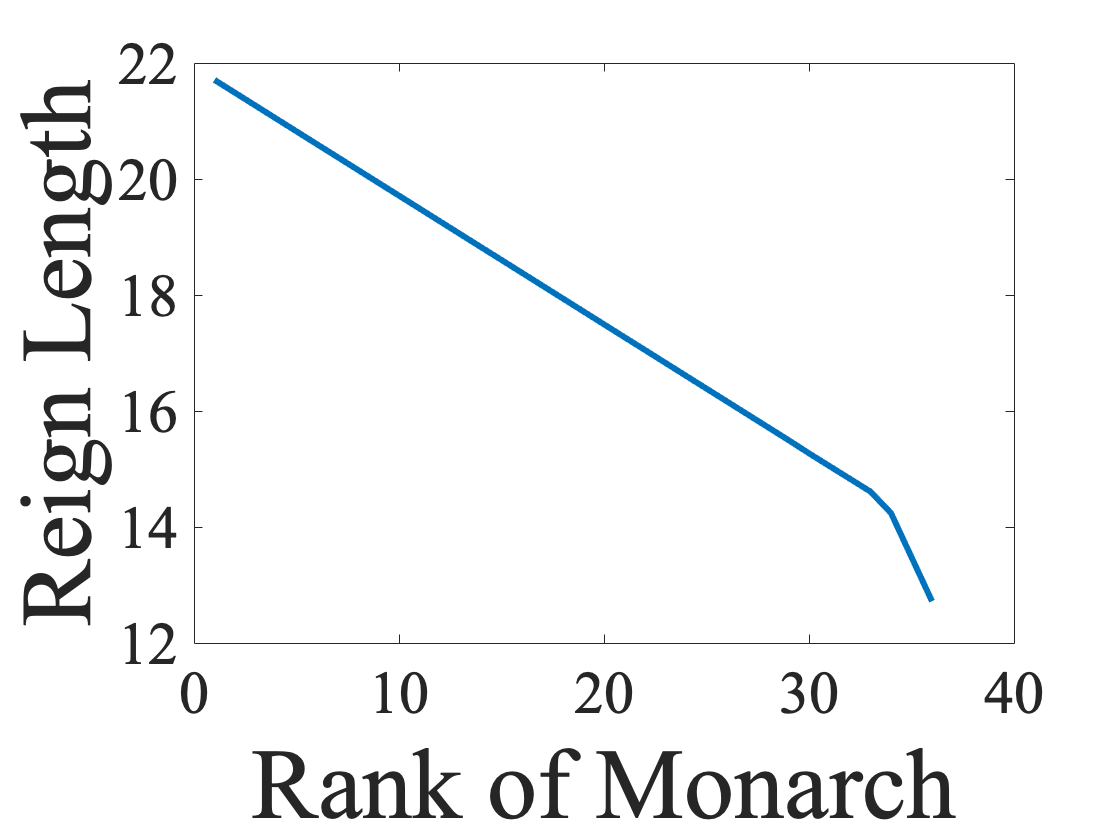} & 
\includegraphics[width=1in]{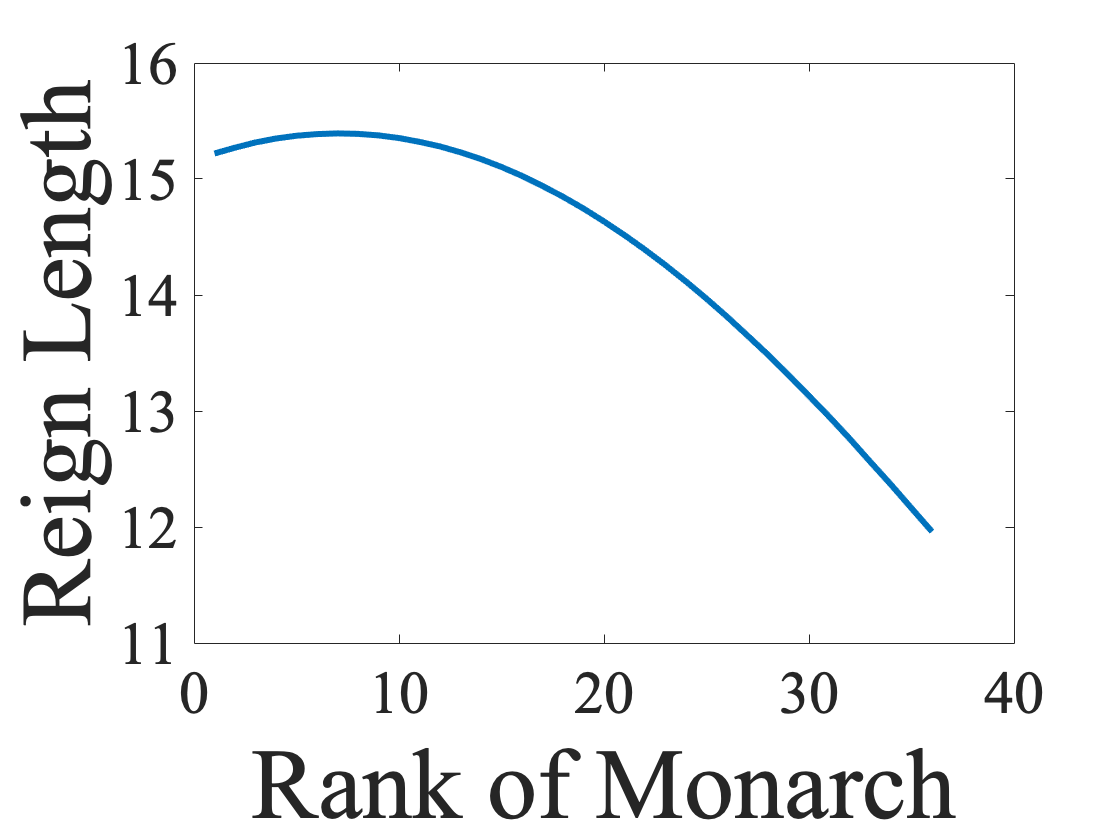} \\
Best GON & Best DNN & Best FICNN & Best GPR   \\
\end{tabular}
\caption{Illustrative Example: Best fits for four methods for the 1D Monarchs' Reigns problem detailed in Sec. \ref{sec:kings}: the goal is to predict the rank of the monarch that will rule longest in a given dynasty. The predicted maximizer of the GON and DNN coincide at the 6th monarch, the convex fit (FICNN) is very rigid and does poorly, whereas the GPR fit is smooth and reasonable.  The GON shown is the composition of the component functions $c(x)$ and $u(\cdot)$ that are shown in Fig. \ref{fig:1dunimodal}.} \label{fig:kings}
\end{figure}

We will define GONs by using the shape constraint \emph{unimodality}: a function is unimodal if it has a maximizer and is non-increasing along any ray that starts at that maximizer (see the GON in Fig. \ref{fig:kings} for an example unimodal function). A few papers have studied learning \emph{1D} unimodal functions \cite{Stout:2008, BSsplines:2014, Dunson:2005, Chatterjee:2019}. This paper goes beyond those prior work both in fitting 1D unimodal functions \emph{without} prior knowledge of the maximizer using constrained empirical risk minimization, and in the ability to fit \emph{multi-D} unimodal functions with a known maximizer. 

Recently, \citet{Gupta:2020} did show how to construct and fit a \emph{subclass} of multi-d unimodal functions by applying linear inequality constraints on the parameters of a lattice model.  However, their unimodality constraints were overly restrictive in that they were separable by dimension, and hence were sufficient but not necessary for multi-d unimodality. We will give a new set of linear inequality constraints that we are \emph{both necessary and sufficient} for a lattice function to be unimodal. This paper also differs from \citet{Gupta:2020} in that we use unimodality to create surrogate functions for global optimization.

\section{Global Optimization Networks}\label{sec:gons}
We propose a new multi-layer function class that we call \emph{global optimization networks} (GONs) defined as a unimodal function composed with an invertible function, and a conditional variant CGONs.

\subsection{Definition of Global Optimization Networks}\label{sec:gonshapeconstraints}
\begin{figure}
\centering
\includegraphics[width=3.25in]{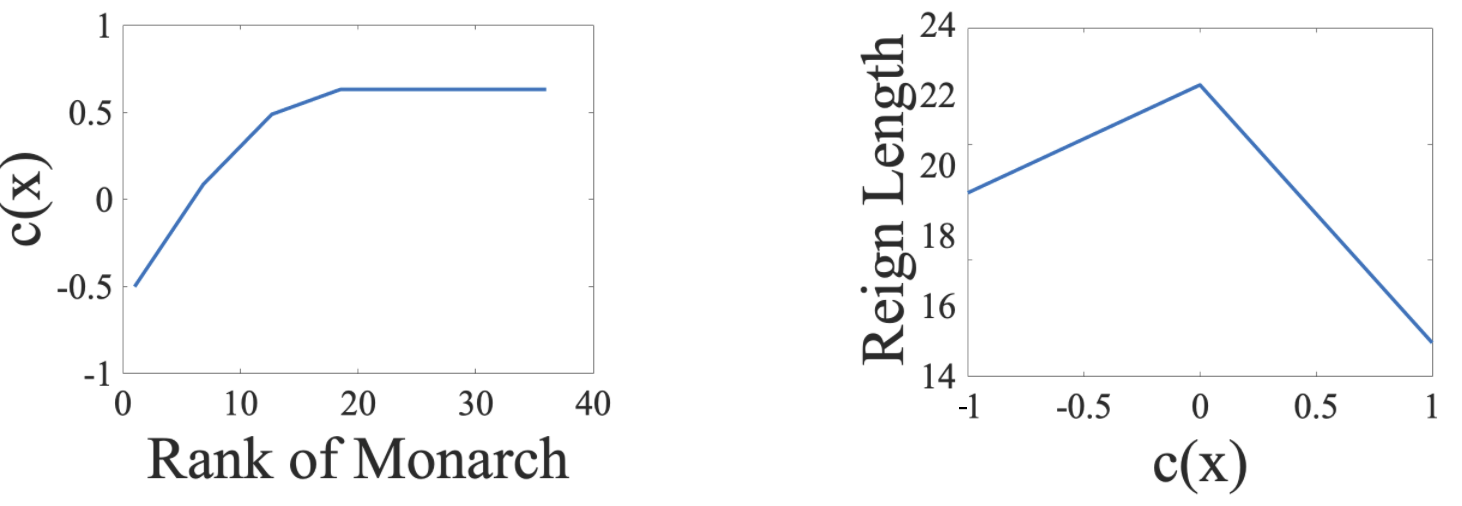}
\caption{The component $c(x)$ and $u(x)$ fit for the best GON for the 1D  Monarchs' Reigns dataset detailed in Sec. \ref{sec:kings}. \textbf{Left:} The first-layer $c(x)$ is a piece-wise linear function (PLF) is defined by five key-value pairs, and meets the invertibility requirement because it is strictly monotonically increasing. \textbf{Middle:} The second-layer $u(\cdot)$ is a PLF defined by three key-value pairs, and was constrained to be unimodal around $0$ by making it monotonically increasing up to $0$ and monotonically decreasing after $0$ and forcing its middle knot to be at $0$, and the other two knots were fixed at $-1$ and $1$ so the 2nd-layer PLF can be described as a 1D lattice function.
The resulting GON $h(x;\bldphi)= u(c(x))$ is shown in Fig. \ref{fig:kings}. As described in Sec. \ref{sec:train}, the parameters of $c$ and $u$ were trained jointly using constrained empirical risk minimization with linear inequality constraints to ensure the needed monotonicity constraints. }\label{fig:1dunimodal}
\end{figure}

We define a GON to be any multi-layer function $h:\mathbb{R}^D \rightarrow \mathbb{R}$ that can be expressed as $h(\bldx;\bldphi) = u(c(\bldx))$, where $c:\RR^D \rightarrow \calS_D$ is any invertible function whose image $\calS_D$ is a convex subset of $\RR^D$ that contains $\bldzero$, and $u:\calS_D \rightarrow \mathbb{R}$ is any \emph{unimodal} function such that it is non-increasing along any ray that starts at $\bldzero$. 
Fig.~\ref{fig:1dunimodal} shows a 1D example of a $c(x)$ and $u(x)$, with the resulting GON $h(\bldx) = u(c(x))$ shown at the far-left of Fig. \ref{fig:kings}. The role of the $c(\bldx)$ is to stretch, rotate, and shift where the outputs of $c$ land in $u$'s domain so that the GON maximizer $\hat{\bldx}$ satisfies $c(\hat{\bldx}) = 0$, which the unimodal function $u$ then maps to the GON maximum. 

Because $\bldzero$ is the maximizer of $u$, we have the GON maximizer $\hat{\bldx} = c^{-1}(\bldzero)$, where $c$ is invertible because it is a bijection. The maximizer $\hat{\bldx} \equiv \arg \max_\bldx h(\bldx;\bldphi) = c^{-1}(\bldzero)$ will be efficient to find if $c$ is efficient to invert at $\bldzero$. Further, suppose $c(\bldx)=s(c'(\bldx))$ for some bijective $c':\RR^D\to\calS_D$ and bijective $s:\calS_D\to\calS_D$ with $s(\bldzero)=\bldzero$. The GON maximizer $\hat{\bldx}  = c^{-1}(\bldzero) = c'^{-1}(s^{-1}(\bldzero)) = c'^{-1}(\bldzero)$, thus only $c'$ must be computationally easy to invert, and $s$ can be quite flexible to increase the expressiveness of $c$.



\subsection{Relation of GONs to Other Function Classes} \label{sec:gonflexibility}
We show how GONs are related to other function classes. All proofs for this paper are in the Appendix. First, note that a unimodal function with an arbitrary maximizer is a special case of a GON:

\begin{restatable}{prop}{PropUnimodalIsGON}\label{prop:PropUnimodalIsGON}
\textbf{Prop. 1:} Let $g:\calS_D\rightarrow \RR$ be a unimodal function with global maximizer $\bldx^* \in\calS_D\subseteq\RR^D$. Then $g$ can be expressed as a GON. 
\end{restatable}

We note concave/convex functions are a special case of unimodal functions (and thus GONs generalize ICNNs \cite{Zico:2017}): 

\begin{restatable}{prop}{PropConvexFunctionIsUnimodal}\label{prop:PropConvexFunctionIsUnimodal}
\textbf{Prop. 2:} Let $g:\calS_D\rightarrow \RR$ be a concave function with global maximizer $\hat{\bldx}\in\calS_D\subseteq\RR^D$. Then $g$ is unimodal with maximizer $\hat{\bldx}$.
\end{restatable}

Surprisingly, continuous \emph{1D} GONs are always unimodal:

\begin{restatable}{prop}{PropOneDGonIsUnimodal} \label{prop:PropOneDGonIsUnimodal}
\textbf{Prop. 3:} Let $u:\calS_1 \rightarrow \mathbb{R},\calS_1\subseteq\RR$ be a 1D unimodal function with maximizer at $0$. Let $c: \mathbb{R} \rightarrow \calS_1$ be continuous, bijective, and have $0$ in its image. Then $h(x;\bldphi) = u(c(x))$ is unimodal.
\end{restatable}

\subsection{Conditional Global Optimizaton Networks}
Consider the conditional global optimization problem: $\bldx^* = \arg \max_\bldx g(\bldx, \bldz)$ for $\bldz\in\RR^M$ \citep{Zico:2017}.  For example, conditioned on what percentage of a job is manual labor $z$, predict the number of weekly work hours $x^*$ that will maximize long-term output \citep{Pencavel:2015}.

We extend the GON definition to a \emph{conditional global optimization network} (CGON). Let $s$ and $c'$ be as defined above, $\bldz\in\RR^M$ be an $M$-dimensional feature vector of conditional inputs, and $r:\mathbb{R}^M \rightarrow \calS_D$ be any learnable function. We define a CGON to be any function that can be written as:
\begin{align}
    h(\bldx,\bldz;\bldphi) = u\left(s\left(\frac{c'\left(\bldx\right) + r\left(\bldz\right)}{2}\right)\right).
\end{align} 
Since the maximizer of $u$ is fixed at $\bldzero$, and $s(\bldzero) = \bldzero$, it is easy to show the CGON maximizer is at $\hat{\bldx} = c'^{-1}(-r(\bldz))$. Note $r$ is unrestricted, so the CGON can have arbitrary dependence on the conditional input $\bldz$, for example $r$ can be a DNN. Table~\ref{tab:buildingblocks} summarizes the requirements for GON and CGON. 

\section{Constructing GONs and CGONs}\label{construction}
Next, we describe how to build GONs by using piecewise linear functions (PLFs) and their multi-d cousins, lattice functions. The benefits of using these function classes are their: \emph{(i)} \textbf{flexibility:} they are respectively universal approximators of continuous bounded 1D and multi-D functions over convex domains; \emph{(ii)} \textbf{efficiency:} we show they enable finding the maximizers of GONs and CGONs in $\mathcal{O}(D)$ time; and \emph{(iii)} \textbf{trainability:} we show they can be trained using constrained empirical risk minimization with linear equality constraints.

\begin{table}
\caption{GON and CGON summaries for $\bldx\in\RR^D$ and $\bldz\in\RR^M$.}
\label{tab:buildingblocks}
\vskip 0.15in
\begin{center}
\begin{small}
\begin{sc}
\begin{tabular}{lll}
\toprule
 & GON & CGON\\
\midrule
Formulation & $h(\bldx;\bldphi)=u(c(\bldx))=u(s(c'(\bldx)))$ & $h(\bldx,\bldz;\bldphi)=u\left(s\left(\frac{c'(\bldx) + r(\bldz)}{2}\right)\right)$ \\
Maximizer   & $\hat{\bldx} = c^{-1}(\bldzero)=c'^{-1}(\bldzero)$  & $\hat{\bldx} = c'^{-1}(-r(\bldz))$ \\
Req. on $c':\RR^D\to\calS_D$   & bijective, easy to invert   & bijective, easy to invert \\
Req. on $s:\calS_D\to\calS_D$   & bijective, $s(\bldzero)=\bldzero$    & bijective, $s(\bldzero)=\bldzero$ \\
Req. on $u:\calS_D\to\RR$   & unimodal, $\arg\max_\bldx u(\bldx)=\bldzero$  & unimodal, $\arg\max_\bldx u(\bldx)=\bldzero$ \\
Req. on $r:\RR^M\to\calS_D$   & -    & any function \\
\bottomrule
\end{tabular}
\end{sc}
\end{small}
\end{center}
\vskip -0.1in
\end{table}

\subsection{Constructing 1D GONs with PLFs}\label{sec:1dgons}
We first show how to construct efficient two-layer GONs using piecewise-linear functions (PLFs) for both the invertible $c()$ and unimodal $u()$, as in Fig. \ref{fig:1dunimodal}. Recall that a PLF can be defined by a set of key-value pairs, and then is evaluated at any point by linearly interpolating the values of the surrounding two keypoints. Let $c(x)$ be defined by the $K^{(c)}$ key-value pairs $(\kappa_k^{(c)} \in \mathbb{R}, \nu_k^{(c)} \in \mathbb{R})$ for $k = 1, \ldots, K^{(c)}$.  Then $c(x) =$
\begin{align}
\sum_{i=1}^{K^{(c)}-1} \left(\nu^{(c)}_i + \frac{x-\kappa^{(c)}_i}{\kappa^{(c)}_{i+1}-\kappa^{(c)}_{i}} \left(\nu^{(c)}_{i+1} - \nu^{(c)}_i\right) \right)I_{\kappa^{(c)}_i < x \leq \kappa^{(c)}_{i+1}}. \label{eq:plf}
\end{align}
In our experiments, we fix the keys of $c$ to be the two endpoints of the feasible input domain plus the $K^{(c)}-2$ quantiles of the inputs in the train data, and only train the PLF values $\{\nu^{(c)}_k\}$.

Recall that a 1D continuous invertible function defined on a closed interval must be strictly monotonic. One can make a PLF monotonically increasing(decreasing) by constraining its values to be increasing(decreasing) (as done in isotonic regression \cite{Barlow:1972}). In addition, we constrain the outputs of $c$ to lie within the input domain of the second-layer function $u$, which we set to be $[-(K^{(u)} - 1)/2, -(K^{(u)} + 1)/2]$ as explained below. Thus the parameters $\{\nu^{(c)}_k\}$ of the PLF $c$ are constrained to satisfy:   
\begin{align}
    -\frac{K^{(u)}-1}{2} \leq \nu_1^{(c)}<\ldots<\nu_{K^{(c)}}^{(c)}\leq \frac{K^{(u)}-1}{2}. \label{eq:rangemonotonic}
\end{align}


To construct a unimodal PLF with maximizer at $0$, use an odd number of $K^{(u)}$ keypoints uniformly spaced in $\calS_1=[-(K^{(u)}-1)/2, (K^{(u)}-1)/2]$. Hence, $\kappa_i=-(K^{(u)}-1)/2+i-1, i=1,\ldots,K^{(u)}$. $K^{(u)}$ is a hyperparameter, where a larger value of $K^{(u)}$ increases the number of parameters of $u$ and hence the flexibility of $u$. Since $K^{(u)}$ must be odd, this makes 0 the middle keypoint of $u$. We then constrain the PLF to be increasing up to 0, and decreasing after 0. That is, a PLF with $K^{(u)}$ keypoints satisfies the unimodality constraints if their values $\nu_k^{(u)}$ satisfies the following $K^{(u)}-1$ linear inequality constraints:
\begin{align}
    \nu_1^{(u)}<\ldots<\nu^{(u)}_{(K^{(u)}+1)/2}>\ldots>\nu^{(u)}_{K^{(u)}}.
\end{align}

Note that the domain of $u$ is bounded by its first and last keypoints, i.e., $\calS_1=[-(K^{(u)}-1)/2, (K^{(u)}-1)/2]$, which is why in (\ref{eq:rangemonotonic}) we constrained the outputs of $c$ to land there.

All continuous 1D functions defined on a closed interval can be approximated arbitrarily well by a PLF. It follows that $g(x)$ can be approximated arbitrarily well by a $h(x;\bldphi)$ constructed with PLFs as well. Thus, this construction can approximate arbitrarily well all continuous 1D GON functions defined on closed intervals.

Given this PLF construction, to find the maximizer of $h(x)$, we only need to invert $c(x)$ at 0. Since $c$ is a monotonically increasing PLF, inverting it is efficient and requires a constant number of operations: first find $c$'s smallest keypoint $\kappa^*$ that satisfies $c(\kappa^*) \geq 0$, and then invert the linear segment between this keypoint and the keypoint to the left of it to get $\hat{x}  = c^{-1}(0)$. Note that such a $\kappa^*$ must exist since we assume that 0 is in the image of $c$.

\subsection{Multi-D GONs Using Lattice Layers}\label{sec:multidgons}
Our multi-D GON construction is a generalization of our 1D construction. For $c$, we simply use $D$ monotonic PLFs, one for each input, which is.a common first layer for deep lattice networks \citep{GuptaEtAl:2016,canini:2016,You:2017}, and constrain their output ranges to the domain of $u$ using linear inequality constraints like \eqref{eq:rangemonotonic}.  One can increase the GON's flexibility by setting $s:\calS_D\to\calS_D$ to be cascades of no-bias hyperbolic-tangent-activated dense layers, or other invertible models \citep{Behrmann:2019}, but our experiments simply use $D$ PLFs for $c$.  

For $u$, we use a $D$-dimensional lattice function \cite{Garcia:2009}, and we propose new linear inequality constraints for a lattice that are both sufficient and necessary to ensure the lattice is unimodal.

\subsection{Lattice Function Review}
Lattice functions are just multi-D look-up tables that are interpolated to form piecewise multilinear polynomial functions; see~\citet{GuptaEtAl:2016} for more details. Let $\bldV\in\NN^D$ be hyperparameter vector where $\bldV[d]$ is the number of keypoints (and hence flexibility) of the lattice function over its $d$th input. The lattice is defined by the set of $\prod_d\bldV[d]$ regularly-spaced keys or vertices,  
\begin{align*}
\M_\bldV=&\left\{-\left\lfloor\frac{\bldV[1]-1}{2}\right\rfloor,\ldots,\left\lceil\frac{\bldV[1]-1}{2}\right\rceil\right\} \times \ldots\times \\
&\left\{-\left\lfloor\frac{\bldV[D]-1}{2}\right\rfloor,\ldots,\left\lceil\frac{\bldV[D]-1}{2}\right\rceil\right\},
\end{align*}
and corresponding $\prod_d\bldV[d]$ values, $\{\bldtheta_\bldv\ :\bldv\in\M_\bldV\}$, where the keys $\M_\bldV$ are pre-determined and fixed, and the values $\{\bldtheta_\bldv\}$ are trained. The domain of the lattice function $u$ is the ``interior'' of $\M_\bldV$ given by 
\begin{align}
\calS_D=&\left[-\left\lfloor\frac{\bldV[1]-1}{2}\right\rfloor,\left\lceil\frac{\bldV[1]-1}{2}\right\rceil\right]\times\ldots\times \nonumber \\
&\left[-\left\lfloor\frac{\bldV[D]-1}{2}\right\rfloor,\left\lceil\frac{\bldV[D]-1}{2}\right\rceil\right]\subset\RR^D. \label{eq:latticedomain}
\end{align}

To evaluate the lattice function $u(\cdot)$, we find the set of $2^D$ vertices surrounding $\bldx$ given by $\N(\bldx){=}
 \big\{{\floor{\bldx[1]}}, {\floor{\bldx[1]}}{+}1\big\}
 \times\ldots\times
 \big\{{\floor{\bldx[D]}}, {\floor{\bldx[D]}}{+}1\big\}$ 
 and linearly interpolate their parameters using standard multilinear interpolation, i.e.,
\begin{equation}
\label{eq:multilinear_lattice}
u(\bldx) = \smashoperator{\sum_{\bldv\in\N(\bldx)}} \bldtheta_\bldv \Phi_\bldv(\bldx),
\end{equation}
where $\Phi_\bldv(\bldx)$ is the linear interpolation weight on vertex $\bldv$ given by
\begin{equation}
\begin{aligned}
\label{eq:phi-def}
\Phi_\bldv(\bldx)=\prod_{d=1}^{D} 
      \left(1+(\bldx[d] - \bldv[d])(-1)^{I_{\bldv[d]=\floor{\bldx[d]}}}\right),
\end{aligned}
\end{equation}
and $I$ is the standard indicator function.

\subsection{Unimodal Lattice Functions}
To make a unimodal lattice, we set each $\bldV[d]$ to be an odd number, and fix the center of the lattice's domain at $\bldzero$. Then, we show one needs the following necessary and sufficient constraints on the lattice parameters for unimodality:

\begin{restatable}{lem}{LemLatticeUnimodalCriteria}
\label{lem:lattice-unimodal-criteria}
\textbf{Lemma 1:} Let $u:\calS_D\rightarrow\RR$ be the function of a $D$-dimensional lattice of size $\bldV\in\NN^D$. For $d=1,\ldots,D$, denote by $\blde_d\in\{0,1\}^D$ the one-hot vector with $\blde_d[i]=1$ iff $i=d$, and for $n\in\NN$, denote by $[n]$ the set $\{1,\ldots,n\}$. Let $s\in[D]$. Every restriction of $u$ to a function with $s$ inputs obtained by fixing the last $D-s$ inputs to constants is unimodal with respect to the maximizer $\bldzero\in\RR^s$ iff for every $\bldv\in\M_\bldV$, $\delta_1,\ldots,\delta_s\in\{0,1\}$  such that $\bldv+\delta_d\blde_d,\bldv-(1-\delta_d)\blde_d\in\M_\bldV$ for all $d\in[s]$ , it holds that
\begin{equation}
    \label{eqn:lattice-unimodality-cond}
    \sum_{d=1}^s(\theta_{\bldv+\delta_d\blde_d}-\theta_{\bldv-(1-\delta_d)\blde_d})\bldv[d] \leq 0.
\end{equation}
\end{restatable}

\subsection{Finding The Maximizer}
Recall that the maximizer of $u$ is at $\bldzero$ by construction, so the maximizer of $h(\bldx;\bldphi)=u(s(c'(\bldx)))$ is $\hat{\bldx} = c'^{-1}(\bldzero)$. Because in our proposed lattice GON construction $c'$ is $D$ PLFs, the $d$th component of the maximizer is found by simply inverting the $d$th PLF of $c'$, which takes $\mathcal{O}(D)$ time overall. 

\subsection{GONs Generalize Unimodality}
Unlike 1D GONs, multi-D GONS generalize unimodal functions:

\begin{restatable}{prop}{PropGONIsNotUnimodal}\label{prop:PropGONIsNotUnimodal}
\textbf{Prop. 4:} Multi-dimensional GONs \emph{generalize} unimodal functions.
\end{restatable}

\subsection{Higher-D GONs with Ensemble of Lattices}
A single unimodal lattice must be defined on a regular grid of at least three keypoints over each feature, thus it needs at least $3^D$ parameters. For better scaling in $D$, we use an ensemble of $T$ lattices \citep{canini:2016} for $u$. 

Let $c:\RR^D\to\calS_D$ be $D$ 1D monotonic PLFs, with $\calS_D=[-V,V]^D$, for some uniform lattice side size $2V+1\in\NN$. We define the ensemble GON as 
\begin{align}
    h(\bldx;\bldphi) = \alpha_0 + \sum_{t=1}^T a_t u_t(\pi_t(c(\bldx))),\label{eq:ensemble}
\end{align} 
where each $\pi_t:\calS_D\rightarrow \calS_Q$ for $t=1,\ldots,T$ with $S_Q=[-V,V]^Q$, is a random projection given by $
\pi_t(\bldx)=\left(\bldx\left[i_{t,1}\right],\bldx\left[i_{t,2}\right],\ldots,\bldx\left[i_{t,Q}\right]\right)$,
and each $u_t(x):\calS_Q\to\RR,\bldzero\in\calS_Q\subseteq\RR^Q$ is a unimodal lattice as described above that acts on a (randomly selected) subset of $Q$ entries of $\bldx$. The $T$ and $Q\leq D$ are hyperparameters; larger $T$ and $Q$ increases the flexibility of the model. The $\alpha_0$ and $\alpha_t \geq 0, t=1,\dots,T$ are learned ensemble parameters.

Prop. 5 shows that the ensemble function in \eqref{eq:ensemble} is still unimodal with maximizer $\bldzero$, and thus one can again find its maximizer by simplying invert the first layer PLFs: $\hat{\bldx} = c^{-1}(\bldzero)$.

\begin{restatable}{prop}{PropEnsembleOfUnimodalsIsUnimodal}
\textbf{Prop. 5:} Let $I\subseteq\RR$ be an interval containing $0$. For an integer $d>0$ denote by $\calS_d$ the Cartesian product $I^d$.
Fix an integer $Q>0$, let $u_t:\calS_Q\rightarrow\RR$, $t=1,\ldots,T$ be unimodal functions with maximizer $\bldzero\in\calS_Q$ and let $\pi_t:\calS_D\rightarrow \calS_Q$ be projections given by $\pi_t(\bldx)=(\bldx[i_{t,1}],\bldx[i_{t,2}],\ldots,\bldx[i_{t,Q}])$. Finally, let $u:\calS_D\rightarrow\RR$, be the ensemble function given by $u(\bldx) = a_0+\sum_{t=1}^T a_t u_t(\pi_t(\bldx)), a_t\geq 0$. Then $u(\bldx)$ is unimodal with maximizer $\bldzero\in\calS_D$.
\end{restatable}

\subsection{CGON Maximizer}  \label{sec:cgonMaximizer}
Similarly, using the above constructions for the CGON layers with $D$ PLFs for $c'$, the CGON global maximizer can also be computed in $\mathcal{O}(D)$ time unless the evaluation of $r(\bldz))$ requires more than $\mathcal{O}(D)$. 

\subsection{Training PLF and Lattice GONs}  \label{sec:train}

Given a standard loss $L$ and a training set $\{\bldx_i, y_i\}$ for $i =1, \ldots, N$, collect the parameters of both $c$ and $u$ into a parameter vector $\bldphi \in \mathbb{R}^p$,  collect all the linear inequality constraints to enforce the monotonicity of $c$ and the unimodality of $u$ into one matrix inequality $A^T \bldphi \geq 0$, then train by solving: 
\begin{equation}
\arg \min_{\bldphi} \sum_{i=1}^N L(h(\bldx_i;\bldphi), y_i) \textrm{ such that }  A^T \bldphi \geq 0. \label{eqn:training}
\end{equation}

Note that $A^T \bldphi \geq 0$ in (\ref{eqn:training}) only forces any monotonic functions in $c$ to be \emph{non-decreasing}, so to force $c$ to be \emph{increasing} for invertibility, if there are any flat segments in any $c$, we simply treat the rightmost key's parameter to be larger. 

To solve (\ref{eqn:training}), we extended the TensorFlow Lattice library \cite{TFLatticeBlogPost}, which already provides PLF layers, lattice layers, and monotonicity constraints, to also support our new joint unimodality constraints, which are now in the open-sourced TensorFlow Lattice library. As recommended in \citet{TFLatticeBlogPost}, we \emph{fixed} the keypoints of $c$ at initialization based on the endpoints and quantiles of the input data, did not train the keypoints of $c$, and we project onto the linear inequality constraints in \eqref{eqn:training} after each batch using 10 steps of Dykstra's projection algorithm \cite{Dykstra:1986}. 



\section{Experiments}\label{sec:experiments}
We compare GONs to DNN's, the convex neural networks \citep{Zico:2017}, and GPR at predicting the maximizer (or minimizer) given the same set of $N$ noisy training samples and only one guess. We start with three real-data problems to build intuition. Then we provide statistically significant comparisons for the problem of selecting the best hyperparameters for five image datasets, and simulations. Table \ref{tab:summary} summarizes the experiments. 

\begin{table}
\caption{Summary of Experiments.}
\label{tab:summary}
\vskip 0.15in
\begin{center}
\begin{small}
\begin{sc}
\begin{tabular}{lrrr}
\toprule
Experiment  & \# of Features & \# Training Samples & \# Test Candidates for $x^*$\\
\midrule
Monarch & 1 & 373 & 2 to 28 \\
Puzzle  & 2 & 36 & 27  \\
Wine &  61 & 84,642 & 24,185 \\
Hyperparameters & 7 & 25 & Infinite\footnote{Theoretically finite, but practically infinite}  \\
Griewank & 4--16 & 100--10,000 & Infinite   \\
Rosenbrock & 4--16 & 100--10,000 & Infinite \\
\bottomrule
\end{tabular}
\end{sc}
\end{small}
\end{center}
\vskip -0.1in
\end{table}

\subsection{Experimental Details}
For each experiment and for each method, we train a set of models with different hyperparameter choices, select the best model according to a validation or cross-validation metric (metric described below), then use the global maximizer of a model trained on the selected hyperparameters as the method's predicted maximizer. 

In practice, given a model $h(x)$, one would predict the maximizer over the entire input domain: $\hat{x} = \arg \max_{x\in\RR^D} h(x)$. This is exactly what we do for our two simulations. However, for our real-data experiments we cannot judge arbitrary predictions, because we do not have the true label for every $x$. Instead, for the real-data experiments, we limit the prediction to the inputs seen in the test set: $\hat{x} = \arg \max_{x\in\mathcal{X}_{\textrm{Test}}} h(x)$, where $\mathcal{X}_{\textrm{Test}}$ is the test set inputs for which we have labels.  

For all experiments, we score each prediction $\hat{x}$ of the maximizer by the true label for $\hat{x}$. 

GPR was trained with sklearn's GPR function. All other models were trained in TensorFlow with Keras layers, and used ADAM \cite{Kingma:2015} with a default learning rate of $.001$ and a batch size of $N$ for $N < 100$, $1000$ for the larger wine experiment in Sec~\ref{sec:wine}, and $100$ otherwise. 
FICNN and PICNN used the formulations in (2) and (3) respectively, from \citet{Zico:2017}. For a CGON with $M$-dimensional conditional inputs, we use $r(z)[j] = \sum_{i=1}^M PLF_i^j(z[i]), j = 1,\dots,D$, where $z[i]$ and $r(z)[i]$ denote the $i$-th entry of $z$ and $r(z)$. For simplicity we use $S(x)=x$, i.e., the identity function. Hyperparameter choices are detailed in the Appendix. For training, labels were scaled to lie in $[0,1]$ to make it easier to specify hyperparameter options. All TensorFlow models were trained to minimize MSE loss. Code for all experiments will be made publicly available. 



\subsection{Predict the Longest-Reigning Monarch}\label{sec:kings}
Predict the rank of the monarch in a royal dynasty that is likely to rule the longest, trained the rank-order of each monarch in a dynasty $x \in [1,36]$, and its label $y$ of how many years the $x$th monarch reigned. Fig. \ref{fig:kings} shows the different validated functions given 373 such training samples from 30 dynasties.  The 1d GON model is unimodal, with its peak at the 6th monarch. The DNN model is less smooth with more peaks and valleys, but agrees with the GON model that the global maxima should be at the 6th monarch.  The convex neural network (FICNN) is over-regularized for this problem, and predicts the first monarch will rule the longest. The GPR model predicts the 7th monarch will rule the longest. See the Appendix for more details and results.

\subsection{Predict the Best Selling Jigsaw Puzzle}\label{sec:puzzles}
We partnered with a jigsaw puzzle company to predict what kind of jigsaw puzzle will sell best. This data has been made publicly available at www.kaggle.com/senzhaogoogle/puzzlesales. Each puzzle is characterized by $D=2$ features: the number of pieces in the puzzle in the range $[79, 1121]$, and the century of the artwork rounded to the nearest century from 1500 to 2000. The non-IID train/validation/test sets had $36/32/27$ puzzles that were new in 2017/2018/2019, each puzzle's label was that year's holiday sales.

We optimized over 8 different hyperparameter choices for each model type (see the Appendix for details), scoring candiate model by the actual sales of the validation-set puzzle it predicted would sell best.  Similarly, the test metric was the actual sales of the test puzzle predicted to have the best sales by the optimized trained model.  

\begin{figure}[]
\centering
\begin{tabular}{c}
\includegraphics[width=3.1in]{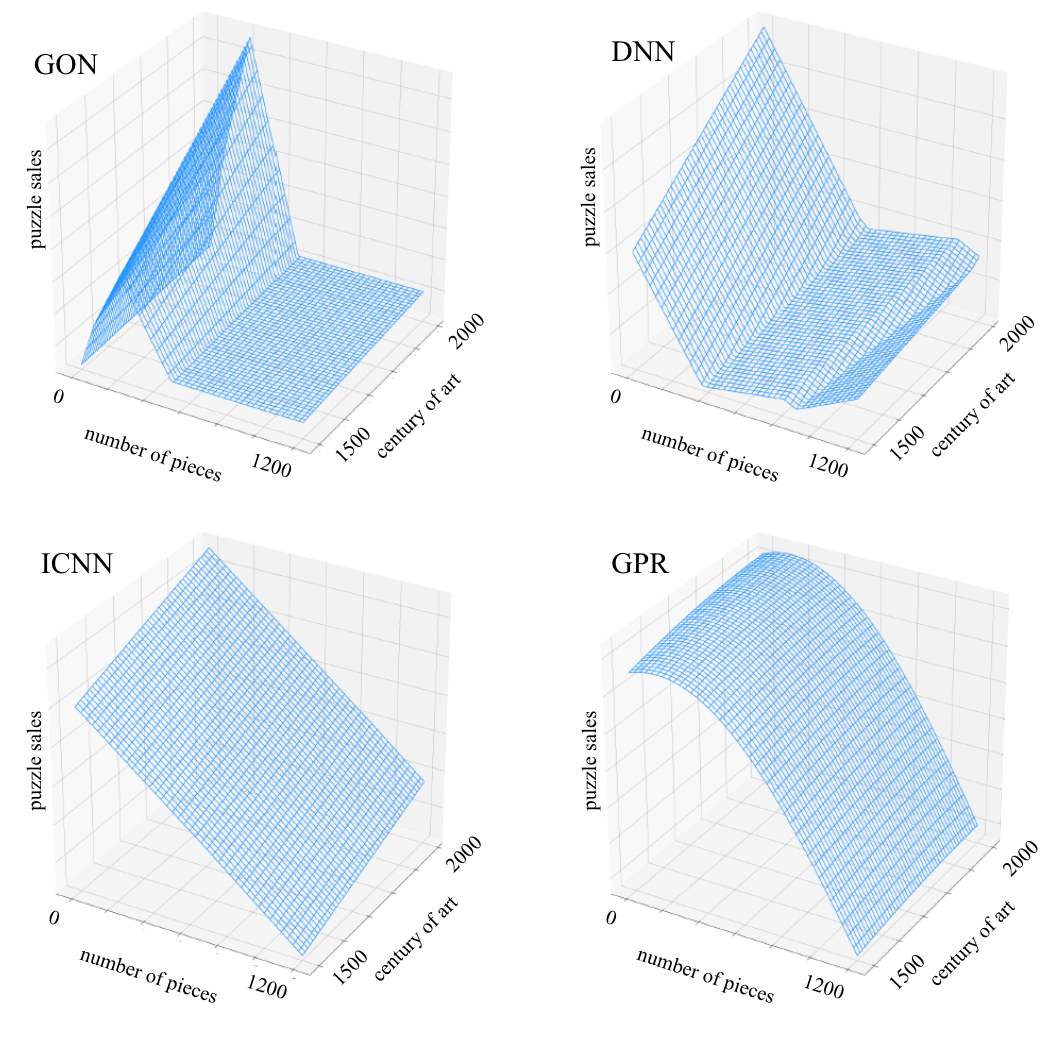}
\end{tabular}
\caption{Validated models for predicting best-selling puzzles. The \emph{global} arg max of the DNN and FICNN predicts the best-selling puzzle would have 0 pieces! That bad extrapolation was fairly stable over hyperparameters (see Appendix).}
\label{fig:bestPuzzles} 
\end{figure}

Figure \ref{fig:bestPuzzles} shows the winning models. Table \ref{tab:puzzles} shows the GON predicted best seller from the test set did have the highest actual sales, and that if one did not restrict the DNN or FICNN to the test set, they predict sales would be maximized by a puzzle with 0 pieces.

\begin{table}[]
\caption{New Puzzle Sales: Metrics for the Trained Models With Best Validation Scores. Bold is best. Train Root MSE and the actual Test Sales (of the test puzzle the surrogate function predicts will sell best) are puzzles sold (scaled). Global Arg Max is the surrogate function's exact global maximizer.}
\label{tab:puzzles}
\vskip 0.15in
\begin{center}
\begin{small}
\begin{sc}
\begin{tabular}{lrrr}
\toprule
& Train & Test& Global Arg Max \\
&RMSE &Sales& \\
& &&\\
\midrule
DNN   & 78.2 &  173 & 0 pieces, year 2000  \\
FICNN    & 78.8 & 173 & 0 pieces, year 2000  \\
GPR    & 87.3 & 2  & 146 pieces, year 2000  \\
GON   & 77.4 & \textbf{182} & 230 pieces, year 2000  \\
\bottomrule
\end{tabular}
\end{sc}
\end{small}
\end{center}
\vskip -0.1in
\end{table}

\subsection{Predict the Highest-Rated Wine}\label{sec:wine}
Using Kaggle data from  Wine Enthusiast Magazine\footnote{www.kaggle.com/dbahri/wine-ratings}, we predict which wine will have the highest quality rating in [80, 100]. We take as given the wine's real-valued price in dollars, 21 Boolean features denoting the country of origin, and 39 Boolean features describing the wine by Wine Enthusiast Magazine for a total of $D=61$ features. There are 84,642 train samples, 12,092 validation samples, and 24,185 test samples, all IID.  We omit results for GPR for this problem because we could not train GPR in sklearn using our machines with 128GB of memory. We validated each model over 15 hyperparameter choices (details in Appendix); the validation score was the actual quality of the model's highest quality prediction over the validation set.  


Table \ref{tab:wine} reports the validated models and their predicted best wines. Consistent across hyperparameter choices, the DNNs and FICNNs relied heavily on the price feature, and the best DNN wrongly predicted that the most expensive test wine would be the highest-quality. 


We also compared the ability of CGON, PICNN, and DNN models to predict the highest-quality wine conditioned on six different price points. We used the same hyperparameters for these models as for the unconditioned experiments. The CGON won or tied 5 of the 6 experiments, and never made an egregious prediction (see Appendix for more). 

\begin{table}[]
\caption{Best Wine: Results for Models With Best Validation Scores. Units are \emph{quality points} from $[80,100]$. Bold is best.}
\label{tab:wine}
\vskip 0.15in
\begin{center}
\begin{small}
\begin{sc}
\begin{tabular}{llll}
\toprule
Model  & Train  & Test  & Predicted \\
&RMSE &Pts& Best Test Wine\\
\midrule
DNN  & 2.54 & 88 & \$3300, acid, juicy, \\ 
& & & tannin, France \\
FICNN   & 2.20 & 94 & \$1100, complex, earth, \\
& & & lees, tight, Austria \\
GON     & 2.28 & \textbf{97} & \$375, acid, bright, \\
& & & complex, elegant,  \\  
&&& refined, structure, \\
&&& tannin, Italy \\
\bottomrule
\end{tabular}
\end{sc}
\end{small}
\end{center}
\vskip -0.1in
\end{table}

\begin{table*}[]
\caption{Mean Test Accuracy $\pm$ 95\% error margin with predicted best hyperparameters. Bold is stat. sig. best or tied for best at 95\% level.}
\label{tab:hparams}
\vskip 0.15in
\begin{center}
\begin{small}
\begin{sc}
\begin{tabular}{llllll}
\toprule
Method   &  CIFAR-10    & CIFAR-100 & Fashion MNIST & MNIST & SVHN  \\
\midrule
GON &\textbf{70.9\%} $\pm$ \textit{0.4\%} &\textbf{37.0\%} $\pm$ \textit{0.4\%} & \textbf{91.1\%} $\pm$ \textit{0.1\%} &\textbf{98.9\%} $\pm$ \textit{0.1\%}& \textbf{87.8}\% $\pm$ \textit{0.3\%}\\
FICNN &67.7\% $\pm$ \textit{1.6\%} & 34.5\% $\pm$ \textit{1.6\%}&\textbf{91.0\%} $\pm$ \textit{0.1\%} &\textbf{99.0\%} $\pm$ \textit{0.1\%}& \textbf{88.3}\% $\pm$ \textit{0.4\%}\\
DNN &67.8\% $\pm$ \textit{1.0\%}& 33.6\% $\pm$ \textit{1.2\%}&90.7\% $\pm$ \textit{0.2\%} &\textbf{99.0\%} $\pm$ \textit{0.1\%}& \textbf{86.1}\% $\pm$ \textit{3.1\%}\\
GPR &66.0\% $\pm$ \textit{3.5\%} & 34.9\% $\pm$ \textit{0.9\%}&90.7\% $\pm$ \textit{0.2\%} &\textbf{98.9\%} $\pm$ \textit{0.1\%}& 85.6\% $\pm$ \textit{1.8\%}\\
\midrule
CGON &\textbf{69.7\%} $\pm$ \textit{0.5\%}&\textbf{35.5\%} $\pm$ \textit{0.6\%}&\textbf{91.3\%} $\pm$ \textit{0.1\%} &\textbf{98.9\%} $\pm$ \textit{0.1\%}& \textbf{87.8}\% $\pm$ \textit{0.3\%}\\
PICNN &65.5\% $\pm$ \textit{3.5\%}& 32.4\% $\pm$ \textit{1.0\%}&91.1\% $\pm$ \textit{0.1\%} &\textbf{99.0\%} $\pm$ \textit{0.1\%}& \textbf{87.3}\% $\pm$ \textit{2.8\%}\\
DNN &67.5\% $\pm$ \textit{1.1\%}&32.5\% $\pm$ \textit{1.7\%}&90.9\% $\pm$ \textit{0.2\%} &\textbf{98.9\%} $\pm$ \textit{0.1\%}& \textbf{87.2}\% $\pm$ \textit{1.4\%}\\
GPR &\textbf{68.1\%} $\pm$ \textit{2.1\%} & \textbf{34.7\%} $\pm$ \textit{0.8\%}&91.1\% $\pm$ \textit{0.1\%} &\textbf{98.9\%} $\pm$ \textit{0.1\%}& \textbf{87.0}\% $\pm$ \textit{1.9\%}\\
\bottomrule
\end{tabular}
\end{sc}
\end{small}
\end{center}
\vskip -0.1in
\end{table*}

\subsection{Hyperparameter Optimization For Image Classifiers} \label{sec:gonhparam}
The next experiment predicts the best hyperparameters for image classifiers. We ran experiments on five benchmark datasets: CIFAR-10/100 \cite{Krizhevsky09learningmultiple}, Fashion MNIST \cite{DBLP:journals/corr/abs-1708-07747}, MNIST \cite{lecun2010mnist} and cropped SVHN \cite{Netzer2011} datasets with their default train/test splits, and use 10\% of the train set as validation. We use ReLU-activated image classifiers: $Conv(f1, k) \to MaxPool(p) \to Conv(f2, k) \to MaxPool(p) \to Conv(f3, k) \to Dense(u) \to Dense(\#classes)$, where filters/units $f1, f2, f3, u \in[8,128]$, kernel/pool size $k,p\in[2,5]$ and training epochs $e\in[1,20]$ are treated as hyperparameters.

To train the optimizers, we randomly sample $N=25$ sets of hyperparameters $(f1,f2,f3,u,k,p,e)$, then train $N=25$ image classifiers on each set of hyperparameters, and use their $N=25$ validation errors as the train labels to fit the response surfaces over the $D=7$ dimensional feature space of hyperparameter choices. For the conditional models, we conditioned on $e=10$ training epochs. 

For GON and CGON, we found the global maximizer of the response surface over the $D=7$ hyperparameter space by inverting the PLFs. For FICNN and PICNN, we used ADAM to find their global maximizers, taking advantage of the fact that their response surfaces are concave, similar to the original work of Zico et al. \cite{Zico:2017}.  For DNN and GPR, we first randomly generated a candidate set $\mathcal{X_{\textrm{candidates}}}$ of 100,000 hyperparameter-sets from the $D=7$-dim domain, and set $\hat{\bldx}=\arg\max_{\bldx\in\mathcal{X_{\textrm{candidates}}}}h(\bldx;\bldphi)$, and use that predicted best hyperparameters to re-train the image classifier and report the test error rates.  For each of the 5 image classification problems, we ran the entire experiment 50 times, each with a different random draw of the $N=25$ random hyperparameters set used to train the response surface. See the Appendix for details. 

Table~\ref{tab:hparams} shows that GON and CGON are statistically significantly the best or tied for the best for all 5 image datasets. 

\subsection{Simulations on Benchmark Functions} \label{sec:gonsims}
We ran extensive simulations with two popular benchmark functions: the banana-shaped Rosenbrock and pocked-convex Griewank functions \cite{Pardalos}.   Table \ref{tab:rosenbrockShort} shows the results for increasing $D$, $N$ and train noise $\sigma$ (see Appendix for full experimental details and results). GON was statistically significantly the best predictor of the global minimizer for all the simulation set-ups for both Rosenbrock (\ref{tab:rosenbrockShort}) and Griewank (Appendix). CGON was also consistently best for Rosenbrock (Appendix). For Griewank, CGON was the best or tied for the best in 6 slices, and PICNN, DNN and GPR were the best or tied for the best in 0, 5 and 3 slices, respectively (Appendix).  


\section{Conclusions}\label{sec:conclusions}
We defined GONs by the shape constraints they must obey: invertible layers and unimodal layers. We showed provide better or comparable accuracy as DNNs, convex functions, and GPR for predicting a maximizer. We focused on using PLF and lattice layers because they are arbitrarily flexible models and amenable to shape constraints, but other invertible layers could be used (e.g. \citet{Behrmann:2019}), or other unimodal (or even convex) layers. Computationally, we found the time to fit a GON was similar to ICNNs and DNNs using Tensorflow for the same number of parameters, but a DLN GON maximizer can be found exactly in $\mathcal{O}(D)$ time. 

\begin{table}[]
\caption{Rosenbrock simulation results with 95\% conf. intervals. Bold is stat. sig. best or tied for best.}
\label{tab:rosenbrockShort}
\vskip 0.15in
\begin{center}
\begin{small}
\begin{sc}
\centering\small
\begin{tabular}{lllll}
\toprule 
$D$ &  GON & FICNN & DNN & GPR \\
\midrule
$4$	&\textbf{213} $\pm$ \textit{24} & 833 $\pm$ \textit{92} & 2259 $\pm$ \textit{151} & 2310 $\pm$ \textit{186} \\
$8$	&\textbf{492} $\pm$ \textit{37} & 2370 $\pm$ \textit{188} & 5019 $\pm$ \textit{209} & 4791 $\pm$ \textit{241} \\ 
$12$  &\textbf{734} $\pm$ \textit{47} & 3575 $\pm$ \textit{278} & 7407 $\pm$ \textit{220} & 7022 $\pm$ \textit{257}  \\
$16$  &\textbf{1004} $\pm$ \textit{21} & 5750 $\pm$ \textit{128} & 9466 $\pm$ \textit{91} & 9133 $\pm$ \textit{107}  \\
\midrule
$\sigma$ &  GON & FICNN & DNN & GPR \\
\midrule
$0.5$    &\textbf{419} $\pm$ \textit{13} & 1273 $\pm$ \textit{52} & 5183 $\pm$ \textit{116} & 3830 $\pm$ \textit{117}  \\
$1.0$    &\textbf{557} $\pm$ \textit{16} & 2805 $\pm$ \textit{97} & 6216 $\pm$ \textit{113} & 5737 $\pm$ \textit{95}  \\
$2.0$    &\textbf{797} $\pm$ \textit{22} & 4382 $\pm$ \textit{118} & 7075 $\pm$ \textit{105} & 6445 $\pm$ \textit{77}  \\
$4.0$    &\textbf{999} $\pm$ \textit{26} & 6383 $\pm$ \textit{139} & 7651 $\pm$ \textit{105} & 6478 $\pm$ \textit{70}  \\
\midrule
$N$ &  GON & FICNN & DNN & GPR \\
\midrule
$1e2$	    &\textbf{473} $\pm$ \textit{9} & 2983 $\pm$ \textit{78} & 6462 $\pm$ \textit{83} & 5820 $\pm$ \textit{90}  \\
$1e3$    &\textbf{897} $\pm$ \textit{19} & 4281 $\pm$ \textit{104} & 6237 $\pm$ \textit{87} & 5923 $\pm$ 97 \\
$1e4$   &\textbf{463} $\pm$ \textit{13} & 2133 $\pm$ \textit{71} & 5414 $\pm$ \textit{97} & 5700 $\pm$ 103 \\
\bottomrule
\end{tabular}
\end{sc}
\end{small}
\end{center}
\vskip -0.1in
\end{table}

\clearpage
\bibliography{references}

\begin{thebibliography}{58}
\providecommand{\natexlab}[1]{#1}
\providecommand{\url}[1]{\texttt{#1}}
\expandafter\ifx\csname urlstyle\endcsname\relax
  \providecommand{\doi}[1]{doi: #1}\else
  \providecommand{\doi}{doi: \begingroup \urlstyle{rm}\Url}\fi

\bibitem[Amos et~al.(2017)Amos, Xu, and Kolter]{Zico:2017}
Amos, B., Xu, L., and Kolter, J.~Z.
\newblock Input convex neural networks.
\newblock In \emph{{ICML}}, pp.\  146--155, 2017.

\bibitem[Archer \& Wang(1993)Archer and Wang]{Archer:93}
Archer, N.~P. and Wang, S.
\newblock Application of the back propagation neural network algorithm with
  monotonicity constraints for two-group classification problems.
\newblock \emph{Decision Sciences}, 24\penalty0 (1):\penalty0 60--75, 1993.

\bibitem[Barlow et~al.(1972)Barlow, Bartholomew, and Bremner]{Barlow:1972}
Barlow, R.~E., Bartholomew, D.~J., and Bremner, J.~M.
\newblock \emph{Statistical inference under order restrictions; the theory and
  application of isotonic regression}.
\newblock Wiley, 1972.

\bibitem[Behrmann et~al.(2019)Behrmann, Grathwohl, Chen, Duvenaud, and
  Jacobsen]{Behrmann:2019}
Behrmann, J., Grathwohl, W., Chen, R. T.~Q., Duvenaud, D., and Jacobsen, J.-H.
\newblock Invertible residual networks.
\newblock In \emph{Proceedings of the 36th International Conference on Machine
  Learning}, pp.\  573--582, 2019.

\bibitem[Bonakdarpour et~al.(2018)Bonakdarpour, Chatterjee, Barber, and
  Lafferty]{Lafferty:2018}
Bonakdarpour, M., Chatterjee, S., Barber, R.~F., and Lafferty, J.
\newblock Prediction rule reshaping.
\newblock 2018.

\bibitem[Box \& Wilson(1951)Box and Wilson]{BoxWilson:1951}
Box, G. E.~P. and Wilson, K.~B.
\newblock On the experimental attainment of optimum conditions.
\newblock \emph{Journal of the Royal Statistical Society: Series B},
  13:\penalty0 1--45, 1951.

\bibitem[Boyle \& Dykstra(1986)Boyle and Dykstra]{Dykstra:1986}
Boyle, J.~P. and Dykstra, R.~L.
\newblock A method for finding projections onto the intersection of convex sets
  in {Hilbert} spaces.
\newblock In \emph{Advances in Order Restricted Statistical Inference}, pp.\
  28--47. 1986.

\bibitem[Canini et~al.(2016)Canini, Cotter, Gupta, Milani~Fard, and
  Pfeifer]{canini:2016}
Canini, K., Cotter, A., Gupta, M., Milani~Fard, M., and Pfeifer, J.
\newblock Fast and flexible monotonic functions with ensembles of lattices.
\newblock In \emph{Advances in Neural Information Processing Systems 29}, pp.\
  2919--2927. 2016.

\bibitem[Cannon(2018)]{Cannon:2018}
Cannon, A.~J.
\newblock Non-crossing nonlinear regression quantiles.
\newblock \emph{Stochastic Environmental Research and Risk Assessment},
  32:\penalty0 3207–3225, 2018.

\bibitem[Chatterjee \& Lafferty(2019)Chatterjee and Lafferty]{Chatterjee:2019}
Chatterjee, S. and Lafferty, J.
\newblock Adaptive risk bounds in unimodal regression.
\newblock \emph{Bernoulli}, 2019.

\bibitem[Chen \& Samworth(2016)Chen and Samworth]{Chen:2016}
Chen, Y. and Samworth, R.~J.
\newblock Generalized additive and index models with shape constraints.
\newblock \emph{Journal Royal Statistical Society B}, 2016.

\bibitem[Chen et~al.(2019)Chen, Shi, and Zhang]{Chen:ICLR2019}
Chen, Y., Shi, Y., and Zhang, B.
\newblock Optimal control via neural networks: A convex approach.
\newblock In \emph{International Conference on Learning Representations}, 2019.

\bibitem[Chen et~al.(2020)Chen, Shi, and Zhang]{Chen:2020}
Chen, Y., Shi, Y., and Zhang, B.
\newblock Data-driven optimal voltage regulation using input convex neural
  networks.
\newblock \emph{Electric Power Systems Research}, 189, 2020.

\bibitem[Chetverikov et~al.(2018)Chetverikov, Santos, and
  Shaikh]{Chetverikov:2018}
Chetverikov, D., Santos, A., and Shaikh, A.
\newblock The econometrics of shape restrictions.
\newblock \emph{Annual Review of Economics}, 10:\penalty0 31--63, 2018.

\bibitem[Cotter et~al.(2019)Cotter, Gupta, Jiang, Louidor, Muller, Narayan,
  Wang, and Zhu]{SetConstraintsICML2019}
Cotter, A., Gupta, M., Jiang, H., Louidor, E., Muller, J., Narayan, T., Wang,
  S., and Zhu, T.
\newblock Shape constraints for set functions.
\newblock In \emph{Proceedings of the 36th International Conference on Machine
  Learning}, pp.\  1388--1396, 2019.

\bibitem[Daniels \& Velikova(2010)Daniels and Velikova]{Daniels:2010}
Daniels, H. and Velikova, M.
\newblock Monotone and partially monotone neural networks.
\newblock \emph{IEEE Transactions on Neural Networks}, 21\penalty0
  (6):\penalty0 906--917, 2010.

\bibitem[Dugas et~al.(2009)Dugas, Bengio, Bélisle, Nadeau, and
  Garcia]{Bengio:2009}
Dugas, C., Bengio, Y., Bélisle, F., Nadeau, C., and Garcia, R.
\newblock Incorporating functional knowledge in neural networks.
\newblock \emph{Journal of Machine Learning Research}, 10\penalty0
  (42):\penalty0 1239--1262, 2009.

\bibitem[Duindam(2015)]{Dynasties:2015}
Duindam, J.
\newblock Dynasties.
\newblock \emph{Medieval Worlds}, 2:\penalty0 59--78, 2015.

\bibitem[Feldman et~al.(2014)Feldman, Gupta, and Frigyik]{Feldman:2014}
Feldman, S., Gupta, M.~R., and Frigyik, B.~A.
\newblock Revisiting {Stein’s} paradox: Multi-task averaging.
\newblock \emph{Journal of Machine Learning Research}, 15\penalty0 (106), 2014.

\bibitem[Garcia \& Gupta(2009)Garcia and Gupta]{Garcia:2009}
Garcia, E. and Gupta, M.
\newblock Lattice regression.
\newblock In \emph{Advances in Neural Information Processing Systems 22}, pp.\
  594--602. 2009.

\bibitem[Garcia et~al.(2012)Garcia, Arora, and Gupta]{Garcia:12}
Garcia, E., Arora, R., and Gupta, M.~R.
\newblock Optimized regression for efficient function evaluation.
\newblock \emph{IEEE Transactions on Image Processing}, 21\penalty0
  (9):\penalty0 4128--4140, 2012.

\bibitem[Gasthaus et~al.(2019)Gasthaus, Benidis, Wang, Rangapuram, Salinas,
  Flunkert, and Januschowski]{gasthaus2019}
Gasthaus, J., Benidis, K., Wang, Y., Rangapuram, S.~S., Salinas, D., Flunkert,
  V., and Januschowski, T.
\newblock Probabilistic forecasting with spline quantile function {RNN}s.
\newblock In \emph{{AIStats}}, volume~89, pp.\  1901--1910, 2019.

\bibitem[Gorissen et~al.(2010)Gorissen, Couckuyt, Demeester, Dhaene, and
  Crombecq]{Gorissen:2010}
Gorissen, D., Couckuyt, I., Demeester, P., Dhaene, T., and Crombecq, K.
\newblock A surrogate modeling and adaptive sampling toolbox for computer based
  design.
\newblock \emph{Journal of Machine Learning Research}, 11\penalty0
  (68):\penalty0 2051--2055, 2010.

\bibitem[Groeneboom \& Jongbloed(2014)Groeneboom and
  Jongbloed]{Groeneboom:2014}
Groeneboom, P. and Jongbloed, G.
\newblock \emph{Nonparametric estimation under shape constraints}.
\newblock Cambridge University Press, 2014.

\bibitem[Gunn \& Dunson(2005)Gunn and Dunson]{Dunson:2005}
Gunn, L.~H. and Dunson, D.~B.
\newblock A transformation approach for incorporating monotone or unimodal
  constraints.
\newblock \emph{Biostatistics}, 6\penalty0 (3):\penalty0 434--449, 2005.

\bibitem[Gupta et~al.(2016)Gupta, Cotter, Pfeifer, Voevodski, Canini, Mangylov,
  Moczydlowski, and van Esbroeck]{GuptaEtAl:2016}
Gupta, M., Cotter, A., Pfeifer, J., Voevodski, K., Canini, K., Mangylov, A.,
  Moczydlowski, W., and van Esbroeck, A.
\newblock Monotonic calibrated interpolated look-up tables.
\newblock \emph{Journal of Machine Learning Research}, 17\penalty0
  (109):\penalty0 1--47, 2016.

\bibitem[Gupta et~al.(2018)Gupta, Bahri, Cotter, and Canini]{Gupta:2018}
Gupta, M., Bahri, D., Cotter, A., and Canini, K.
\newblock Diminishing returns shape constraints for interpretability and
  regularization.
\newblock In \emph{Advances in Neural Information Processing Systems 31}, pp.\
  6834--6844. 2018.

\bibitem[Gupta et~al.(2020)Gupta, Louidor, Mangylov, Morioka, Narayan, and
  Zhao]{Gupta:2020}
Gupta, M.~R., Louidor, E., Mangylov, O., Morioka, N., Narayan, T., and Zhao, S.
\newblock Multidimensional shape constraints.
\newblock In \emph{Proceedings of the 37th International Conference on Machine
  Learning}, 2020.

\bibitem[Horst \& Pardalos(1995)Horst and Pardalos]{Pardalos}
Horst, R. and Pardalos, P.~M.
\newblock \emph{Handbook of Global Optimization}.
\newblock Springer, 1995.

\bibitem[Howard \& Jebara(2008)Howard and Jebara]{Jebara:2007}
Howard, A. and Jebara, T.
\newblock Learning monotonic transformations for classification.
\newblock In \emph{Advances in Neural Information Processing Systems 20}, pp.\
  681--688. 2008.

\bibitem[Jones(2001)]{Jones:2001}
Jones, D.~R.
\newblock A taxonomy of global optimization methods based on response surfaces.
\newblock \emph{Journal of Global Optimization}, 21:\penalty0 345--383, 2001.

\bibitem[Kennedy \& Eberhart(1995)Kennedy and Eberhart]{Kennedy:1995}
Kennedy, J. and Eberhart, R.
\newblock Particle swarm optimization.
\newblock In \emph{Proceedings of {IEEE} International Conference on Neural
  Networks}, pp.\  1942--1948, 1995.

\bibitem[Kim et~al.(2004)Kim, Lee, Vandenberghe, and Yang]{Kim:2004}
Kim, J., Lee, J., Vandenberghe, L., and Yang, C.-K.~K.
\newblock Techniques for improving the accuracy of geometric-programming based
  analog circuit design optimization.
\newblock In \emph{IEEE/ACM International Conference on Computer Aided Design},
  pp.\  863--870, 2004.

\bibitem[Kingma \& Ba(2015)Kingma and Ba]{Kingma:2015}
Kingma, D.~P. and Ba, J.
\newblock Adam: {A} method for stochastic optimization.
\newblock In \emph{International Conference on Learning Representations}, 2015.

\bibitem[K\"{o}llmann et~al.(2014)K\"{o}llmann, Bornkamp, and
  Ickstadt]{BSsplines:2014}
K\"{o}llmann, C., Bornkamp, B., and Ickstadt, K.
\newblock Unimodal regression using {Bernstein–Schoenberg} splines and
  penalties.
\newblock \emph{Biometrics}, 70\penalty0 (4):\penalty0 783--793, 2014.

\bibitem[Krizhevsky(2009)]{Krizhevsky09learningmultiple}
Krizhevsky, A.
\newblock Learning multiple layers of features from tiny images.
\newblock Technical report, 2009.

\bibitem[LeCun et~al.(2010)LeCun, Cortes, and Burges]{lecun2010mnist}
LeCun, Y., Cortes, C., and Burges, C.
\newblock {MNIST} handwritten digit database.
\newblock \emph{ATT Labs [Online]. Available:
  http://yann.lecun.com/exdb/mnist}, 2, 2010.

\bibitem[Magnani \& Boyd(2009)Magnani and Boyd]{Boyd:2009}
Magnani, A. and Boyd, S.~P.
\newblock Convex piecewise-linear fitting.
\newblock \emph{Optimization and Engineering}, 10\penalty0 (1):\penalty0 1--17,
  2009.

\bibitem[Milani~Fard(2020)]{TFLatticeBlogPost}
Milani~Fard, M.
\newblock {TensorFlow Lattice: Flexible, Controlled, and Interpretable ML},
  2020.

\bibitem[Minin et~al.(2010)Minin, Velikova, Lang, and Daniels]{Minin:2010}
Minin, A., Velikova, M., Lang, B., and Daniels, H.
\newblock Comparison of universal approximators incorporating partial
  monotonicity by structure.
\newblock \emph{Neural Networks}, 23\penalty0 (4):\penalty0 471--475, 2010.

\bibitem[Netzer et~al.(2011)Netzer, Wang, Coates, Bissacco, Wu, and
  Ng]{Netzer2011}
Netzer, Y., Wang, T., Coates, A., Bissacco, A., Wu, B., and Ng, A.~Y.
\newblock Reading digits in natural images with unsupervised feature learning.
\newblock 2011.

\bibitem[Nocedal \& Wright(2006)Nocedal and Wright]{NocedalBook}
Nocedal, J. and Wright, S.~J.
\newblock \emph{Numerical Optimization}.
\newblock Springer, 2006.

\bibitem[Pei \& Hu(2018)Pei and Hu]{PeiHu:2018}
Pei, S. and Hu, Q.
\newblock Partially monotonic decision trees.
\newblock \emph{Information Sciences}, 424:\penalty0 104--117, 2018.

\bibitem[Pencavel(2015)]{Pencavel:2015}
Pencavel, J.
\newblock The productivity of work hours.
\newblock \emph{The Economic Journal}, 125:\penalty0 2052--2076, 2015.

\bibitem[Pinker(2011)]{BetterAngels}
Pinker, S.
\newblock \emph{The Better Angels Of Our Nature: Why Violence Has Declined}.
\newblock Viking Penguin, 2011.

\bibitem[Pya \& Wood(2015)Pya and Wood]{Pya:2015}
Pya, N. and Wood, S.~N.
\newblock Shape constrained additive models.
\newblock \emph{Statistics and Computing}, 2015.

\bibitem[Qian et~al.(2015)Qian, Xu, Liang, Liu, and Wang]{Qian:2015}
Qian, Y., Xu, H., Liang, J., Liu, B., and Wang, J.
\newblock Fusing monotonic decision trees.
\newblock \emph{IEEE Transactions on Knowledge and Data Engineering},
  27\penalty0 (10):\penalty0 2717--2728, 2015.

\bibitem[Qu \& Hu(2011)Qu and Hu]{QuHu:11}
Qu, Y.-J. and Hu, B.-G.
\newblock Generalized constraint neural network regression model subject to
  linear priors.
\newblock \emph{IEEE Transactions on Neural Networks}, 22\penalty0
  (11):\penalty0 2447--2459, 2011.

\bibitem[Rasmussen \& Williams(2006)Rasmussen and Williams]{GPRBook}
Rasmussen, C.~E. and Williams, C. K.~I.
\newblock \emph{Gaussian Processes for Machine Learning}.
\newblock MIT Press, 2006.

\bibitem[Shi \& Eberhart(1998)Shi and Eberhart]{Shi:1998}
Shi, Y. and Eberhart, R.
\newblock A modified particle swarm optimizer.
\newblock In \emph{Proceedings of {IEEE} International Conference on
  Evolutionary Computation}, pp.\  69--73, 1998.

\bibitem[Sill(1998)]{Sill:98}
Sill, J.
\newblock Monotonic networks.
\newblock In \emph{Advances in Neural Information Processing Systems 10}, pp.\
  661--667. 1998.

\bibitem[Stout(2008)]{Stout:2008}
Stout, Q.~F.
\newblock Unimodal regression via prefix isotonic regression.
\newblock \emph{Computational Statistics and Data Analysis}, 53:\penalty0
  289--297, 2008.

\bibitem[Wang \& Gupta(2020)Wang and Gupta]{WangGupta:2020}
Wang, S. and Gupta, M.~R.
\newblock Deontological ethics by monotonicity shape constraints.
\newblock In \emph{AIStats}, 2020.

\bibitem[Wehenkel \& Louppe(2019)Wehenkel and Louppe]{Louppe:2019}
Wehenkel, A. and Louppe, G.
\newblock Unconstrained monotonic neural networks.
\newblock \emph{Advances in Neural Information Processing Systems}, 2019.

\bibitem[Xiao et~al.(2017)Xiao, Rasul, and
  Vollgraf]{DBLP:journals/corr/abs-1708-07747}
Xiao, H., Rasul, K., and Vollgraf, R.
\newblock Fashion-{MNIST}: a novel image dataset for benchmarking machine
  learning algorithms.
\newblock \emph{CoRR}, abs/1708.07747, 2017.

\bibitem[You et~al.(2017)You, Ding, Canini, Pfeifer, and Gupta]{You:2017}
You, S., Ding, D., Canini, K., Pfeifer, J., and Gupta, M.
\newblock Deep lattice networks and partial monotonic functions.
\newblock In \emph{Advances in Neural Information Processing Systems 30}, pp.\
  2981--2989. 2017.

\bibitem[Zhang \& Zhang(1999)Zhang and Zhang]{ZhangZhang:1999}
Zhang, H. and Zhang, Z.
\newblock Feedforward networks with monotone constraints.
\newblock In \emph{International Joint Conference on Neural Networks},
  volume~3, pp.\  1820--1823, 1999.

\bibitem[Zhu et~al.(2017)Zhu, Tsanga, Wang, and Ashfaq]{Zhu:2017}
Zhu, H., Tsanga, E. C.~C., Wang, X.-Z., and Ashfaq, R. A.~R.
\newblock Monotonic classification extreme learning machine.
\newblock \emph{Neurocomputing}, 225:\penalty0 205--213, 2017.

\end{thebibliography}
\bibliographystyle{icml2022}

\clearpage
\appendix
\onecolumn

\section{Broader Related Work}
In the next two subsections, we review the broader set of related work for GONs: strategies in fitting functions for optimization, and shape constraints. 

\subsection{Related Work in Fitting Functions for Optimization}
The idea of fitting a function and then predicting the maximizer to be the maximizer of the fitted-function (see Fig. 1) dates back to at least Box and Wilson's 1951 paper \cite{BoxWilson:1951}, which considered fitting interpolating high-order polynomials through all the data (but in practice restricted their experiments to linear and quadratic functions). Such fits are often called \emph{response surfaces} or \emph{surrogates}. This strategy is also used as an intermediary step for convex optimization in \emph{trust region methods} that fit a quadratic function locally to a neighborhood, and then expand or contract the region over which the quadratic is fitted \cite{NocedalBook}.  They also considered two issues we do not address in this paper. First, they considered the selection of training examples that would lead to good estimates, e.g., by properly covering the input space, whereas in this paper we take the training examples as given. Second, they noted that one might need to fit a series of such surrogate functions over different subregions of the input space, and we leave this question of specifying a good multi-pass global optimization algorithm open for future work.

\citet{Zico:2017} proposed fitting flexible convex (or concave) functions to all the training data. They constructed convex functions through a multi-layer ReLU-activated machine-learned model with the appropriate monotonicity shape constraints to get the convexity. They proposed a \emph{fully-input convex neural network} (referred to as FICNN or just ICNN) for solving the global optimization problem $x^* = \arg \min_\bldx g(\bldx)$, and a \emph{partial-input convex neural network} (PICNN) for the \emph{conditional} global optimization problem $\bldx^* = \arg \min_\bldx g(\bldx, \bldz)$. Because their machine-learned functions $h(\bldx;\bldphi)$ and $h(\bldx,\bldz;\bldphi)$ are convex in $\bldx$, they can be minimized numerically to find $\arg\min_\bldx h(\bldx;\bldphi)$ and $\arg\min_\bldx h(\bldx,\bldz;\bldphi)$.  Others have found this strategy useful  \citep{Chen:ICLR2019,Chen:2020}. However, note that ReLU-activated ICNNs are neither smooth nor strongly convex, which reduces the convergence rate in finding the minimizer of an ICNN.

For non-convex problems, \cite{Jones:2001} contended that fitting quadratics is ``unreliable'' because ``the surface may not sufficiently capture the shape of the function.'' Arbitrary machine-learning models have been used as surrogate models \cite{Gorissen:2010}. However, for those methods, we cannot use gradient-based methodologies to find their maximizers, and hence the second stage of finding the global optimizer of such models becomes computationally restrictive in high-dimensions. In addition, using an arbitrary surrogate misses the chance to semantically regularize the fitted function to have a shape with a unique global optimum. 

A different flexible fitting strategy is kriging, also called Gaussian process regression (GPR) \cite{GPRBook}. GPR \emph{interpolates} the training set \cite{Jones:2001}. Computing GPR has complexity $O(N^3)$ for $N$ training examples, and finding its optimizer is problematic as the number of inputs $D$ increases \cite{Jones:2001,GPRBook}. 

Compared to the prior work, the proposed GON functions are more flexible than concave functions, but do have a unique global maximizer. Further, the  global maximizer of GONs can be specified analytically and found in $\mathcal{O}(D)$ time, without the need for gradient-based algorithms. Further, unlike methods which use arbitrarily flexible fits like DNNs, the proposed GONs use a semantically meaningful regularization strategy, which produces more interpretable and often more accurate results, as shown in Sec. \ref{sec:experiments}.

\subsection{Related Work in Shape Constraints}
\begin{figure}[!]
\centering
\includegraphics[width=3.5in]{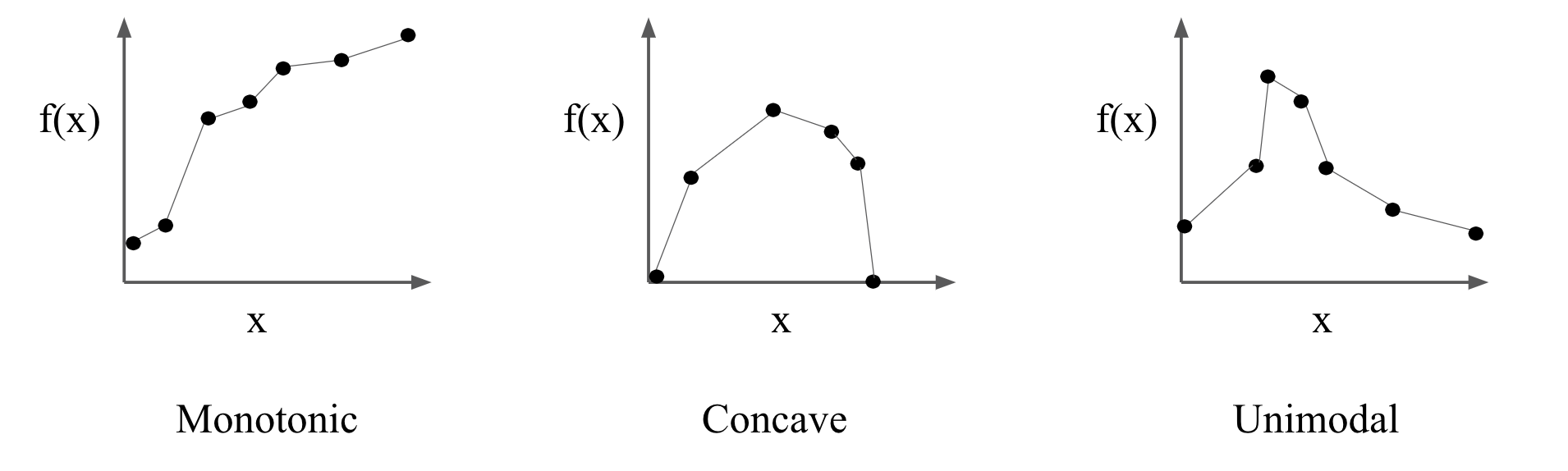}
\caption{Examples of piece-wise linear functions (PLFs) that satisfy different shape constraints. Each PLF is parameterized by the key-value pairs marked by the black dots.}\label{fig:shapeConstraintExamples}
\end{figure}

\emph{Shape constraints} define function classes by specifying their model shape properties \cite{Groeneboom:2014, Chetverikov:2018}. Fig. \ref{fig:shapeConstraintExamples} shows some 1D examples. 

The most common and popular shape constraint is \emph{monotonicity}. For a 1D function with $x \in \mathbb{R}$, a function is monotonically increasing if $f(x)$ is non-decreasing as $x$ increases, or monotonically decreasing if the opposite.  Here we use the shorthand \emph{monotonic} for either direction. For differentiable functions, a function is monotonic if the first derivative is non-negative everywhere.  An example of a simple 1D monotonic function class is the set of linear functions with positive slopes. 

A popular flexible 1D function class for satisfying shape constraints is piecewise linear functions (PLF) \cite{Barlow:1972, Jebara:2007, Groeneboom:2014, Garcia:12, GuptaEtAl:2016}, as shown in Figure \ref{fig:shapeConstraintExamples}.

Monotonicity constraints can also be applied to multi-dimensional functions with $\bldx \in \mathbb{R}^D$, where the usual definition is that $f(\bldx)$ is increasing in the $d$th feature, $\bldx[d]$, if $f(\bldx)$ is non-decreasing as $\bldx[d]$ increases, with all other features held fixed.  A function can be monotonic with respect to a subset of its features. Flexible multi-dimensional  monotonic functions have been created by constraining  neural networks, \cite[e.g.,][]{Archer:93,Sill:98,ZhangZhang:1999,Daniels:2010,Minin:2010,QuHu:11, Zhu:2017,Cannon:2018,Louppe:2019}, support vector machines \cite{Jebara:2007}, decision trees \cite[e.g.,][]{Qian:2015,PeiHu:2018}, and lattices \cite[e.g.,][]{GuptaEtAl:2016,canini:2016,You:2017}, or by post-processing \cite[e.g.,][]{Chernozhukov:2010,Lafferty:2018}.

Other shape constraints that have been used for machine-learning include diminishing returns \cite{Pya:2015,Chen:2016,Gupta:2018}, complementary inputs \cite{Gupta:2020}, and dominance between inputs \cite{Gupta:2020}.

\begin{figure}[!t]
\centering
\includegraphics[width=4in]{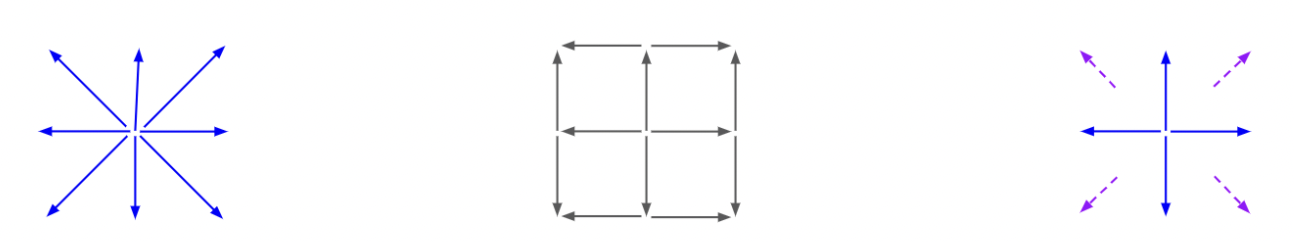}
\caption{Arrows illustrate different unimodality constraints for a two-dimensional function, with the maximizer at the center of each plot. \textbf{Left:} A function is defined to be \emph{unimodal} if it has a maximizer and is non-increasing along all rays starting at the maximizer. \textbf{Middle:} The arrows show the prior art: separable unimodality constraints given in \cite{Gupta:2020} on a $3 \times 3$ grid of knots that is later bilinearly-interpolated to form the lattice function. Each arrow signifies that the parameter value at the arrow tip must be smaller than or equal to the parameter value at its tail.  These separable constraints are sufficient but not necessary for unimodality, as they un-necessarily enforce unimodality on every orthogonal slice of the function.  \textbf{Right:} The joint unimodality constraints proposed in this paper in \eqref{eqn:lattice-unimodality-cond} for a 3 $\times$ 3 lattice. The solid blue arrows indicate that the parameter value at the arrow tip must be smaller than or equal to the parameter value at the arrow tail. The dashed purple arrows signify that the parameter value at the purple tip must be smaller than or equal to the average of the two knot values at the diagonal corners. This set of constraints in \eqref{eqn:lattice-unimodality-cond} is shown to be both sufficient and necessary for a lattice function to be unimodal.}
\label{fig:arrowsPlot} 
\end{figure}

Another overly-restrictive special case of unimodality is jointly concave functions. These have been produced by summing jointly concave basis functions \cite{Kim:2004,Boyd:2009}, or by DNN's with ReLU activations that are constrained to be jointly convex over a subset of features \cite{Bengio:2009,Zico:2017}.  We show experimentally that concave functions are generally too restrictive for finding and understanding global maximizers. 

Shape constraints are often applied to lattice functions, as we do in this paper. Lattices are linearly-interpolated multidimensional look-up tables \cite{Garcia:12}: in one-dimension a lattice is just a piecewise linear function with regular knots. Lattices are arbitrarily flexible, just add more knots (parameters). Because the lattice is parameterized by a regular grid of function values, many shape constraints turn into sparse linear inequality constraints, making training them easy \cite{Gupta:2016,Gupta:2018,Gupta:2020}. Higher-dimensional lattice functions are achieved through ensembles \cite{canini:2016} and multi-layer models \cite{You:2017,SetConstraintsICML2019}. In the next section, we will show how to construct efficient GONs using multi-layer lattice models with the appropriate shape constraints.  Tensorflow Lattice provides an open source library for lattice functions \cite{TFLatticeBlogPost}, we provide extensions to the Tensorflow Lattice library for GONs. 

\section{Block Diagrams for Ensemble GON and CGON Models}
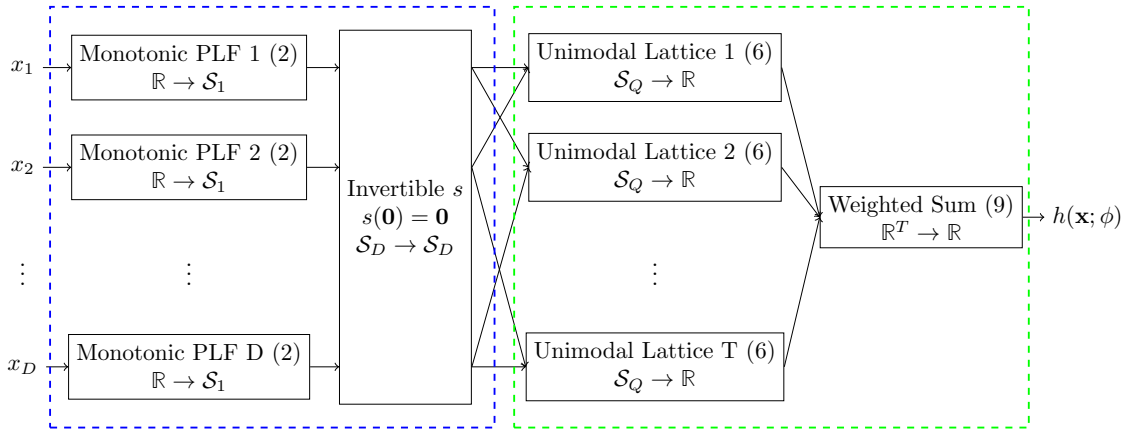
\begin{figure*}[!t]
\centering
\resizebox{6in}{!}{
\begin{tikzpicture}
    \node (dots) at (-2.5,1.5) [align=center] {$\vdots$};
    \node (x_d) at (-2.5,0) [align=center] {$x_D$};
    \node (x_2) at (-2.5,3) [align=center] {$x_2$};    
    \node (x_1) at (-2.5,4.5) [align=center] {$x_1$};

    \node (dots) at (0,1.5) [align=center] {$\vdots$};
    \node (plf_d) at (0,0) [draw, align=center, minimum width=3em] {Monotonic PLF D \eqref{eq:plf} \\ $\RR\to\calS_1$};
    \node (plf_2) at (0,3) [draw, align=center, minimum width=3em] {Monotonic PLF 2 \eqref{eq:plf} \\ $\RR\to\calS_1$};    
    \node (plf_1) at (0,4.5) [draw, align=center, minimum width=3em] {Monotonic PLF 1 \eqref{eq:plf} \\ $\RR\to\calS_1$};

    \draw[->] (x_d.east) --  (plf_d.west);
    \draw[->] (x_2.east) --  (plf_2.west);
    \draw[->] (x_1.east) --  (plf_1.west);
    
    \node (dots) at (7,1.5) [align=center] {$\vdots$};
    \node (u_t) at (7,0) [draw, align=center, minimum width=3em] {Unimodal Lattice T \eqref{eq:multilinear_lattice} \\ $\calS_Q\to\RR$};
    \node (u_2) at (7,3) [draw, align=center, minimum width=3em] {Unimodal Lattice 2 \eqref{eq:multilinear_lattice} \\ $\calS_Q\to\RR$};  
    \node (u_1) at (7,4.5) [draw, align=center, minimum width=3em] {Unimodal Lattice 1 \eqref{eq:multilinear_lattice} \\ $\calS_Q\to\RR$};

    \node (s) at (3.25,2.25) [draw, align=center, minimum width=3em, minimum height=16em] {Invertible $s$ \\ $s(\bldzero)=\bldzero$ \\ $\calS_D\to\calS_D$};

    \node [coordinate] at ($(s.west)+(0, -2.25)$) (s_lower_left) {};
    \node [coordinate] at ($(s.west)+(0, 0.75)$) (s_middle_left) {};
    \node [coordinate] at ($(s.west)+(0, 2.25)$) (s_upper_left) {};
    \node [coordinate] at ($(s.east)+(0, -2.25)$) (s_lower_right) {};
    \node [coordinate] at ($(s.east)+(0, 0.75)$) (s_middle_right) {};
    \node [coordinate] at ($(s.east)+(0, 2.25)$) (s_upper_right) {};

    \draw[->] (plf_d.east) -- (s_lower_left);
    \draw[->] (plf_2.east) -- (s_middle_left);
    \draw[->] (plf_1.east) -- (s_upper_left);
    \draw[->] (s_lower_right) -- (u_t.west);
    \draw[->] (s_lower_right) -- (u_2.west);
    \draw[->] (s_middle_right) -- (u_1.west);
    \draw[->] (s_middle_right) -- (u_t.west);
    \draw[->] (s_upper_right) -- (u_1.west);
    \draw[->] (s_upper_right) -- (u_2.west);

    \node (ensemble) at (11,2.25) [draw, align=center, minimum width=3em] {Weighted Sum \eqref{eq:ensemble} \\ $\RR^{T}\to\RR$};

    \draw[->] (u_t.east) -- (ensemble.west);
    \draw[->] (u_2.east) -- (ensemble.west);
    \draw[->] (u_1.east) -- (ensemble.west);

    \node (output) at (13.5,2.25) [align=center] {$h(\bldx;\bldphi)$};

    \draw[->] (ensemble.east) -- (output.west);

    \node (invertible) at (1.25,2.25) [draw=blue, thick, dashed, minimum width=19em, minimum height=18em] {};
    \node (unimodal) at (8.75,2.25) [draw=green, thick, dashed, minimum width=22em, minimum height=18em] {};

\end{tikzpicture}
}
\caption{Block diagram for the proposed multi-dim GON using a PLF layer for $c(x)$ and an ensemble of weighted unimodal lattices for $u(\cdot)$. Each unimodal lattice $u(\cdot)$ takes a subset of features as the input. The blue box denotes the invertible function and the green box denotes the unimodal function.}
\label{fig:GONconstruction}
\end{figure*}

Fig. \ref{fig:GONconstruction} gives a block diagram for a DLN GON using an ensemble of lattices as a layer (for more on lattice ensembles see \citet{canini:2016}, and for more on ensembles of lattices as a layer in a multi-layer model see \citet{You:2017}. 

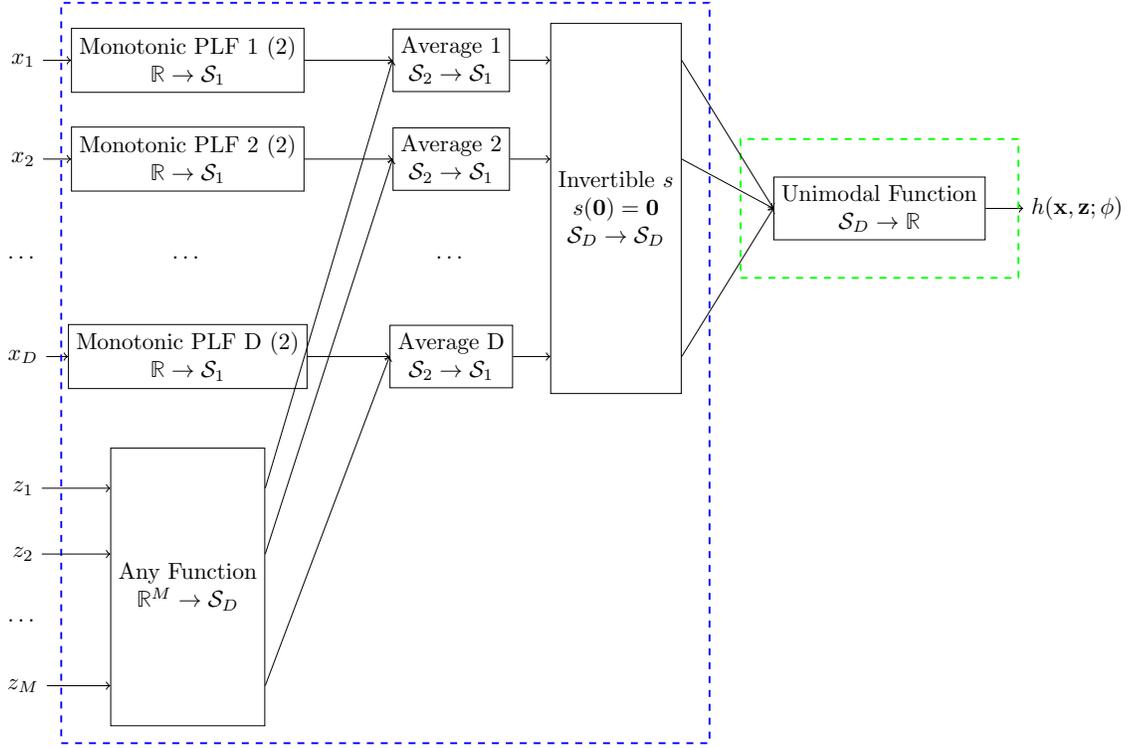
\begin{figure*}[!t]
\centering
\resizebox{6in}{!}{
\begin{tikzpicture}

    \node (dots) at (-2.5,1.5) [align=center] {$\dots$};
    \node (x_d) at (-2.5,0) [align=center] {$x_D$};
    \node (x_2) at (-2.5,3) [align=center] {$x_2$};    
    \node (x_1) at (-2.5,4.5) [align=center] {$x_1$};

    \node (dots) at (0,1.5) [align=center] {$\dots$};
    \node (plf_d) at (0,0) [draw, align=center, minimum width=3em] {Monotonic PLF D \eqref{eq:plf} \\ $\RR\to\calS_1$};
    \node (plf_2) at (0,3) [draw, align=center, minimum width=3em] {Monotonic PLF 2 \eqref{eq:plf} \\ $\RR\to\calS_1$};    
    \node (plf_1) at (0,4.5) [draw, align=center, minimum width=3em] {Monotonic PLF 1 \eqref{eq:plf} \\ $\RR\to\calS_1$};

    \draw[->] (x_d.east) --  (plf_d.west);
    \draw[->] (x_2.east) --  (plf_2.west);
    \draw[->] (x_1.east) --  (plf_1.west);

    \node (dots) at (4,1.5) [align=center] {$\dots$};
    \node (sum_d) at (4,0) [draw, align=center, minimum width=3em] {Average D \\ $\calS_2\to\calS_1$};
    \node (sum_2) at (4,3) [draw, align=center, minimum width=3em] {Average 2 \\ $\calS_2\to\calS_1$};  
    \node (sum_1) at (4,4.5) [draw, align=center, minimum width=3em] {Average 1 \\ $\calS_2\to\calS_1$};

    \draw[->] (plf_d.east) -- (sum_d.west);
    \draw[->] (plf_2.east) -- (sum_2.west);
    \draw[->] (plf_1.east) -- (sum_1.west);

    \node (ensemble) at (10.5,2.25) [draw, align=center, minimum width=3em] {Unimodal Function \\ $\calS_D\to\RR$};

    \node (output) at (13.5,2.25) [align=center] {$h(\bldx,\bldz;\bldphi)$};

    \draw[->] (ensemble.east) -- (output.west);

    \node (invertible) at (3,-0.25) [draw=blue, thick, dashed, minimum width=28em, minimum height=32em] {};
    \node (unimodal) at (10.5,2.25) [draw=green, thick, dashed, minimum width=12em, minimum height=6em] {};

    \node (dots) at (-2.5,-4) [align=center] {$\dots$};
    \node (z_m) at (-2.5,-5) [align=center] {$z_M$};
    \node (z_2) at (-2.5,-3) [align=center] {$z_2$};    
    \node (z_1) at (-2.5,-2) [align=center] {$z_1$};

    \node (r) at (0,-3.5) [draw, align=center, minimum width=3em, minimum height=12em] {Any Function \\ $\RR^M\to\calS_D$};
    \node [coordinate] at ($(r.west)+(0, -1.5)$) (r_lower_left) {};
    \node [coordinate] at ($(r.west)+(0, 0.5)$) (r_middle_left) {};
    \node [coordinate] at ($(r.west)+(0, 1.5)$) (r_upper_left) {};
    \node [coordinate] at ($(r.east)+(0, -1.5)$) (r_lower_right) {};
    \node [coordinate] at ($(r.east)+(0, 0.5)$) (r_middle_right) {};
    \node [coordinate] at ($(r.east)+(0, 1.5)$) (r_upper_right) {};

    \draw[->] (z_m.east) -- (r_lower_left);
    \draw[->] (z_2.east) -- (r_middle_left);
    \draw[->] (z_1.east) -- (r_upper_left);
    \draw[->] (r_lower_right) -- (sum_d.west);
    \draw[->] (r_middle_right) -- (sum_2.west);
    \draw[->] (r_upper_right) -- (sum_1.west);

    \node (s) at (6.5,2.25) [draw, align=center, minimum width=3em, minimum height=16em] {Invertible $s$ \\ $s(\bldzero)=\bldzero$ \\ $\calS_D\to\calS_D$};

    \node [coordinate] at ($(s.west)+(0, -2.25)$) (s_lower_left) {};
    \node [coordinate] at ($(s.west)+(0, 0.75)$) (s_middle_left) {};
    \node [coordinate] at ($(s.west)+(0, 2.25)$) (s_upper_left) {};
    \node [coordinate] at ($(s.east)+(0, -2.25)$) (s_lower_right) {};
    \node [coordinate] at ($(s.east)+(0, 0.75)$) (s_middle_right) {};
    \node [coordinate] at ($(s.east)+(0, 2.25)$) (s_upper_right) {};

    \draw[->] (sum_d.east) -- (s_lower_left);
    \draw[->] (sum_2.east) -- (s_middle_left);
    \draw[->] (sum_1.east) -- (s_upper_left);
    \draw[->] (s_lower_right) -- (ensemble.west);
    \draw[->] (s_middle_right) -- (ensemble.west);
    \draw[->] (s_upper_right) -- (ensemble.west);

\end{tikzpicture}
}
\caption{Block diagram of a CGON constructed with DLN layers. The blue box marks the invertible function, and the green box marks the unimodal function and is identical to the green box in Figure~\ref{fig:GONconstruction}. There are $D$ inputs $\bldx$ to optimize over given values for the $M$ conditional inputs $\bldz$. The model uses $D$ one-dimensional monotonic PLFs to calibrate each of the $D$ inputs $\bldx$, and uses any function (e.g. a DNN) to map the $M$ conditional inputs $\bldz$ to $D$ outputs. We add each of the $D$ calibrated $\bldx$ to one of the $D$ outputs of $\bldz$, resulting in $D$ inputs $s$ to the unimodal function. Note the whole model will be jointly trained, so the $D$ outputs of $r(\bldz)$ will be optimized to be linearly combined with the $c'(\bldx)$. Then the $D$ outputs from $s$ are separated into an ensemble of unimodal lattices as in \eqref{eq:ensemble}, whose outputs are linearly combined to get the final prediction.}
\label{fig:cgon} 
\end{figure*}

Fig. \ref{fig:cgon} gives a block diagram for a DLN CGON.

\section{Puzzles Experiment More Details}
\begin{figure*}[!t]
\centering
\begin{tabular}{c}
\includegraphics[width=5.0in]{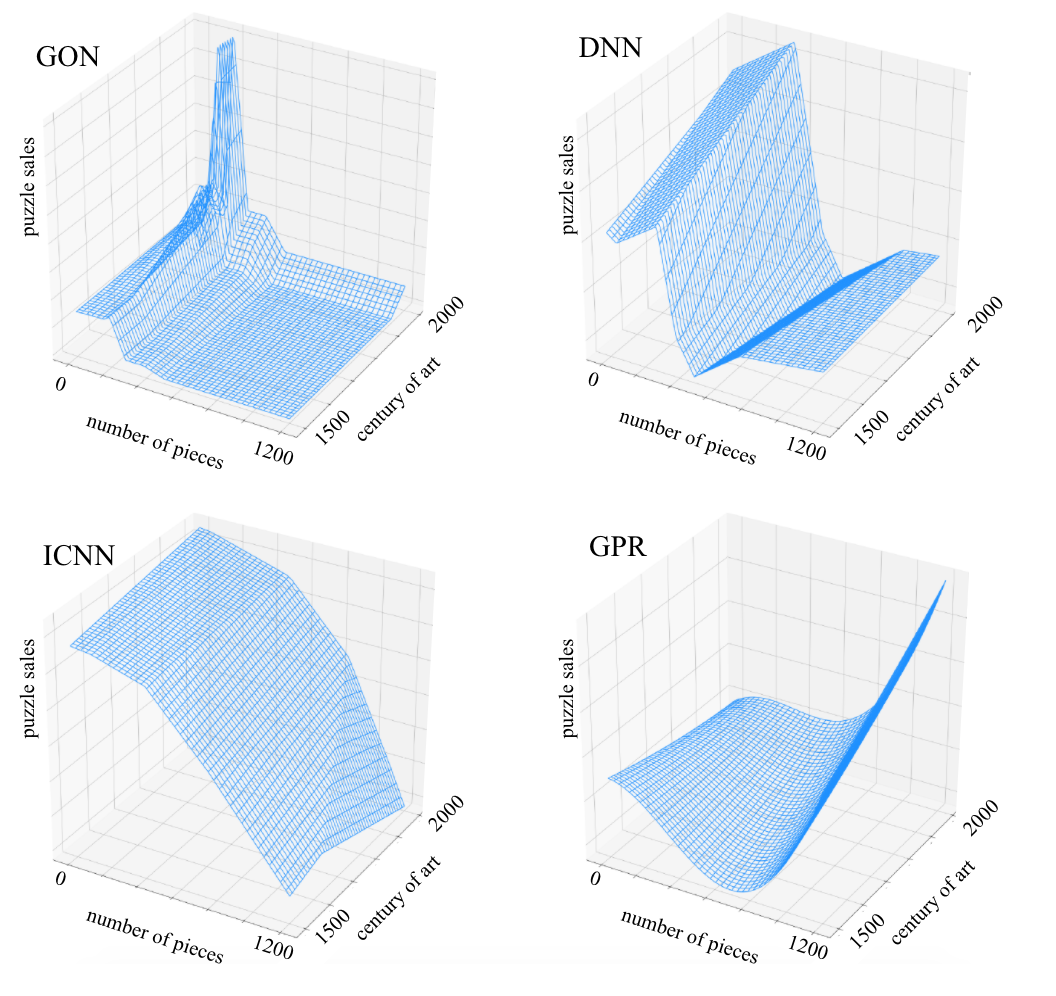}
\end{tabular}
\caption{The most flexible models considered when validating hyperparameters for the puzzle sales experiment. }
\label{fig:puzzles} 
\end{figure*}

To further build intuition, in Figure \ref{fig:puzzles} we show the trained functions with the most flexible hyperparameter choices we validated over. The most flexible GON model used $9$ keypoints for the PLF for each of the two inputs for $c$, and then a $3 \times 3$ lattice for $u$. It has a steep peak at 213 pieces and year 2000. The most flexible DNN, with 4 layers and 8 hidden nodes, is a reasonable model with a peak at 353 pieces and art from year 2000.  The GPR model with $\alpha = 1e-6$ overfit good sales data for one of the largest puzzles.  The most flexible ICNN model, with 4 layers and 8 hidden nodes, still advises the company to make puzzles with zero pieces. 

\section{Proofs}
\begin{figure}[!t]
\centering
\includegraphics[width=4.5in]{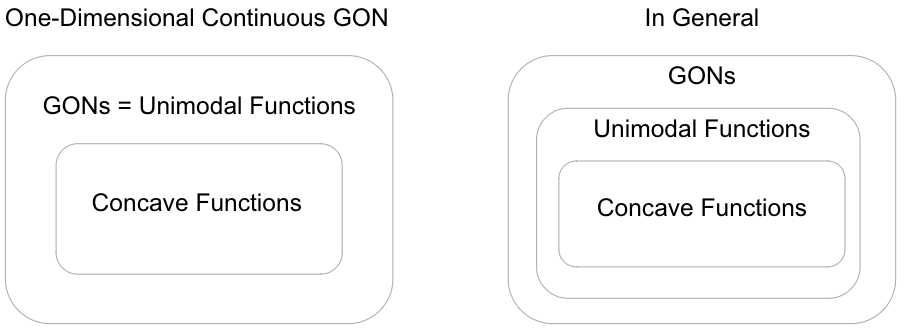}
\caption{Relationship of GONs to other function classes, summarizing Propositions 1, 2, 3, and 4. \textbf{Left:} In the special case of an one-dimensional input, the class of continuous GONs and the class of unimodal functions are identical. \textbf{Right:} For multi-dimensional input spaces, the set of GONs is more expressive than the set of unimodal functions, and unimodal functions more expressive than the set of concave functions.} \label{fig:expressiveness}
\end{figure}

Below are the proofs for all of the results in the paper. See Fig. \ref{fig:expressiveness} for a Venn diagram summarizing Propositions 1,2,3 and 4. 

\subsection{Unimodal Functions Are GONs}
\PropUnimodalIsGON

\begin{proof}
All unimodal functions are GONs. This is because for any unimodal function $g$ with maximizer $\hat{\bldx}$, we can reparametrize it as $g(\bldx)=u(\bldx+\hat{\bldx})$ for some  $u$ that is unimodal with $\arg\max_\bldx u(\bldx)=\bldzero$. This can then be written $u(c(x))$ where $c(\bldx)=\bldx + \hat{\bldx}$ is invertible, thus forming a GON.
\end{proof}

\subsection{Concave Functions Are Unimodal}
\PropConvexFunctionIsUnimodal*
\begin{proof}
Let $r(t)=\hat{\bldx}+t\bldv$, $t\geq0$, be a ray in $\RR^D$ originating at $\hat{\bldx}$. To prove concave $g$ is unimodal, we need to show that $g(r(t))$ is decreasing. Let $r(t_1),r(t_2)$ be two points on the ray with $0\leq t_1 \leq t_2$. Then it's easily verified that $r(t_1)=((t_2-t_1)/t_2) \hat{\bldx} + (t_1/t_2) r(t_2)$. Now by the concavity of $g$, we have
\begin{equation*}
\begin{split}
    g\left(r\left(t_1\right)\right) &= g\left(\frac{t_2-t_1}{t_2} \hat{\bldx} + \frac{t_1}{t_2} r(t_2)\right) \\
    &\geq \frac{t_2-t_1}{t_2} g\left(\hat{\bldx}\right) + \frac{t_1}{t_2} u\left(r\left(t_2\right)\right) \\
    &\geq \frac{t_2-t_1}{t_2} g\left(r\left(t_2\right)\right) + \frac{t_1}{t_2} g\left(r\left(t_2\right)\right) \\
    & = g(r(t_2)),
\end{split}
\end{equation*}
where the last inequality follows since $g(\hat{\bldx})$ is the maximum of $g$.
\end{proof}

\subsection{One-Dimensional GONs With Monotonic $c$ Are Unimodal}
We show that for one-dimensional GONs because a continuous one-to-one function $c$ defined on a convex set must be monotonic, the GON is unimodal.  

\PropOneDGonIsUnimodal
\begin{proof}
Let $x^*\in\RR$ be the pre-image of $0$ under $c$. Consider any $x_1, x_2 \in \RR$ such that  $x_1 < x_2 \leq x^*$.
Note that to be bijective, a continous one-dimensional $c$ with a convex domain must be either monotonically increasing or monotonically decreasing: otherwise, one can find 3 points $x<y<z$ for which either $f(x)<f(y)>f(x)$ or $f(x)>f(y)<f(z)$ and by the mean value theorem it follows that any point $c$ in between $f(x)$ and $f(y)$ will have at least 2 distinct pre-images, contradicting $f$ being one-to-one.  Without loss of generality, assume $c$ is monotonically increasing.  Then we have $c(x_1)\leq c(x_2) \leq c(x^*) = 0$. Since $u$ is unimodal w.r.t its input $0$, we have $u(c(x_1)) \leq u(c(x_2)) \iff {h(x_1) \leq h(x_2)}$. Therefore $h$ is increasing for $x\leq x^*$. An analogous argument shows that $h$ is decreasing for $x \geq x^*$. Thus $h$ is unimodal with respect to $x^*$.
\end{proof}

\subsection{Proof for Linear Inequality Constraints To Make A Lattice Function Unimodal}\label{sec:constraintsProof}

Some visual intution for this lemma is given in Figure \ref{fig:arrowsPlot}.

Throughout this section we use the following notation. For a function $u:\RR^D\rightarrow\RR$ we denote its partial derivative with respect to the $i$th input variable by $\partial_i u$. If $u$ is univariate we denote its derivative by $u'$. For $n\in\NN$ we use $[n]$ to denote the set $\{1,2,\ldots,n\}$ and for $\bldx \in \mathbb{R}^D$, and $d\in[D]$, we denote by $\bldx[d]$ the $d$th entry of $\bldx$. Finally, we denote by $\blde_d \in [0,1]^D$ the one-hot vector where $\blde_d[i] = 1$ iff $i=d$.

Consider a lattice with dimension $D$, size vector $\bldV$, and parameters $\{\theta_\bldv\}_{\bldv\in\M_\bldV}$. For $\bldx\in\RR^D$, we define the cell of $\bldx$ to be the set of its $2^D$ neighboring grid vertices given by $\N(\bldx){=}
\big\{{\floor{\bldx[1]}}, {\floor{\bldx[1]}}{+}1\big\}
\times\ldots\times
\big\{{\floor{\bldx[D]}}, {\floor{\bldx[D]}}{+}1\big\}
$. Then the lattice function $u$ is given by
\begin{equation}
\label{eqn:multilinear_lattice}
u(\bldx) = \smashoperator{\sum_{\bldv\in\N(\bldx)}} \theta_\bldv \Phi_\bldv(\bldx),
\end{equation}
where $\Phi_\bldv(\bldx)$ is the linear interpolation weight on vertex $\bldv$ given by:
\begin{equation}
\begin{aligned}
\label{eqn:phi-def}
\Phi_\bldv(\bldx)=\prod_{d=1}^{D} 
      \Big(1+(\bldx[d] - \bldv[d])(-1)^{I_{\bldv[d]=\floor{\bldx[d]}}}\Big),
\end{aligned}
\end{equation}
and $I$ is the standard indicator function. See~\cite{GuptaEtAl:2016} for more details.

To prove Lemma 1, we'll need the following supporting lemma (Lemma 2), which gives a formula for the partial derivative of a lattice function.

\textbf{Lemma 2:} Let $f:\RR^D\rightarrow\RR$ be a lattice function with dimension $D$, size vector $\bldV$ and parameters $\{\theta_\bldv\}_{\bldv\in\M_\bldV}$.
Then for all $d\in[D]$, and $\bldx\in\M_\bldv$ with $\bldx[d]\not\in\ZZ$ (i.e. $\bldx$ does not lie on the boundary of two adjacent lattice cells in the $d$th direction)
\begin{equation*}
\partial_d f(\bldx)=\smashsum{\bldv\in\N(\bldx)}\Phi_\bldv(\bldx)(\theta_{\ceil{\bldv}_{d,\bldx}}-\theta_{\floor{\bldv}_{d,\bldx}}),
\end{equation*}
where ${\ceil{\bldv}_{d,\bldx}}$ is $\bldv{+}\blde_d$, if $\bldv[d]{=}{\floor{\bldx[d]}}$, or $\bldv$, otherwise, and ${\floor{\bldv}}_{d,\bldx}{=}{\ceil{\bldv}}_{d,\bldx}{-}\blde_d$.

\begin{proof}
Let $\bldx$ satisfy the requirements of the lemma. By~(\ref{eqn:multilinear_lattice}), $\partial_d f(\bldx){=}
\sum_{\bldv\in\N(\bldx)}\theta_\bldv\partial_d \Phi_{\bldv}(\bldx)$.
Denoting by $\lambda(v,x){=}1{+}(x{-}v)(-1)^{I_{v=\floor{x}}}$, for $x{\in}\RR$ and $v{\in}\NN$, we get
\begin{align*}
    \partial_d f(\bldx) &= \smashsum{\bldv\in\N(\bldx)}\theta_\bldv\partial_d \prod_{i=1}^{D}\lambda(\bldv[i],\bldx[i]) \\
    &=\smashsum{\bldv\in\N(\bldx)}\theta_\bldv(-1)^{I_{\bldv[d]=\floor{\bldx[d]}}}\prod_{i\neq d}\lambda(\bldv[i],\bldx[i]),
\end{align*}
where we used the fact that for $x\in\RR\setminus\ZZ$, $\partial \lambda/\partial x=(-1)^{I_{v=\floor{x}}}$.
Partitioning the set $\N(\bldx)$ of size $2^{D}$ into the $2^{D-1}$ pairs $\{(\bldv,\ceil{\bldv}_{d,\bldx}):\bldv\in\N(\bldx),\bldv=\floor{\bldv}_{d,\bldx}\}$, we may regroup the summands to obtain
\begin{align}\label{eqn:partial_d_lattice}
    \partial_d f(\bldx) &=
    \smashsum{\substack{\bldv\in\N(\bldx)\\\bldv=\floor{\bldv}_{d,\bldx}}}
    \big(\theta_{\ceil{\bldv}_{d,\bldx}}-
    \theta_{\floor{\bldv}_{d,\bldx}}\big)
    \prod_{i\neq d}\lambda(\bldv[i],\bldx[i])
\end{align}
Now, observe that $1=\lambda(\floor{\bldx[d]},\bldx[d]) + \lambda(\floor{\bldx[d]}{+}1, \bldx[d])$. Thus, for $\bldv\in\N(\bldx)$ with $\bldv=\floor{\bldv}_{d,\bldx}$, it holds that
\begin{align}\label{eqn:prod_i_neq_d_lambda}
    \nonumber \prod_{i\neq d}\lambda(\bldv[i],\bldx[i]) &= 
    \big(\lambda(\floor{\bldx[d]},\bldx[d]) + \lambda(\floor{\bldx[d]}{+}1, \bldx[d])\big)\cdot\\
\nonumber &\phantom{=}\quad\prod_{i\neq d}\lambda(\bldv[i],\bldx[i]) \\
    &=\Phi_{\bldv}(\bldx)+\Phi_{\ceil{\bldv}_{d,\bldx}}(\bldx).
\end{align}
Substituting~(\ref{eqn:prod_i_neq_d_lambda}) into~(\ref{eqn:partial_d_lattice}), we get
\begin{align*}
    \partial_d f(\bldx) &=
    \smashsum{\substack{\bldv\in\N(\bldx)\\\bldv=\floor{\bldv}_{d,\bldx}}}
    \big(\theta_{\ceil{\bldv}_{d,\bldx}}-
    \theta_{\floor{\bldv}_{d,\bldx}}\big)
    (\Phi_{\bldv}(\bldx)+\Phi_{\ceil{\bldv}_{d,\bldx}}(\bldx)) \\
    &=\smashsum{\bldv\in\N(\bldx)}\Phi_{\bldv}(\bldx)\big(\theta_{\ceil{\bldv}_{d,\bldx}}-
    \theta_{\floor{\bldv}_{d,\bldx}}\big).
\end{align*}
\end{proof}

We are now ready to prove Lemma 1.

\LemLatticeUnimodalCriteria

\begin{proof} 
Every restriction obtained from $u$ by fixing the last $D-s$ features to constants is unimodal w.r.t $\bldzero$ if and only if every such restriction is decreasing along rays originating in $\bldzero$. The latter statement can be equivalently restated as: for each $\bldx\in\RR^D$, the function $f_\bldx:[0,1]\rightarrow\RR$, given by $f_\bldx(t)=u(\bldr_\bldx(t))$, with $\bldr_\bldx(t)=(t\bldx[1],\ldots,t\bldx[s],\bldx[s+1],\bldx[s+2],\ldots,\bldx[D])$, is decreasing. 
Since each such $f_\bldx$ is continuous and piecewise-differentiable with finitely many pieces, the last condition is equivalent to requiring that $f'_\bldx(t) \leq 0$ for all $t \in \interval[open left]{0}{1}$ where the derivative is defined. 
Observe that it's sufficient to require that for all $\bldx\in\RR^D$, $f'_\bldx(1)\leq 0$, when it's defined, since $f'_\bldx(t)=f'_{\bldr_\bldx(t)}(1)/t$. Therefore, statement 1 of the lemma holds if and only if 
\begin{equation}
\label{eqn:directional_deriv>=0}
\forall \bldx\in\RR^D, f'_\bldx(1)\leq 0
\end{equation}
By the chain rule, $f'_\bldx(1){=}\sum_{d=1}^{s}\partial_d u(\bldx){\cdot}\bldx[d]$ and hence using Lemma~\ref{lem:lattice-partial-derivative} we have
\begin{align*}
f'_\bldx(1) 
& = \smashoperator{\sum_{d\in[s],\bldv\in\N(\bldx)}}
\Phi_\bldv(\bldx)(\theta_{\ceil{\bldv}_{d,\bldx}}-\theta_{\floor{\bldv}_{d,\bldx}})
\bldx[d]
\\
& = \smashoperator{\sum_{d,\bldv}}
\Phi_\bldv(\bldx)(\theta_{\ceil{\bldv}_{d,\bldx}}{-}\theta_{\floor{\bldv}_{d,\bldx}})(\bldx[d]{-}\floor{\bldx[d]})\\
&\quad+\smashoperator{\sum_{d,\bldv}}\Phi_\bldv(\bldx)(\theta_{\ceil{\bldv}_{d,\bldx}}{-}\theta_{\floor{\bldv}_{d,\bldx}})\floor{\bldx[d]},
\end{align*}
where to get the last equality we added to and subtracted from each summand the quantity $\Phi_\bldv(\bldx)(\theta_{\ceil{\bldv}_{d,\bldx}}-\theta_{\floor{\bldv}_{d,\bldx}})\floor{\bldx[d]}$.

Next, for a fixed $d\in[D]$, partitioning the set $\N(\bldx)$ of size $2^D$ into the $2^{D-1}$ pairs $\{(\bldv,\floor{\bldv}_{d,\bldx}):\bldv\in\N(\bldx), \bldv=\ceil{\bldv}_{d,\bldx}\}$, we regroup the terms in the summation and get
\begin{align}\label{eqn:unimodality-f'}
\nonumber f'_\bldx(1)
= &\smashoperator{\sum_{d,\bldv:\bldv=\ceil{\bldv}_{d,\bldx}}}
\Big((\Phi_{\bldv}(\bldx){+}\Phi_{\floor{\bldv}_{d,\bldx}}(\bldx))
(\bldx[d]{-}\floor{\bldx[d]}) \\
\phantom{=}& \hspace{2.5em}\times(\theta_{\ceil{\bldv}_{d,\bldx}}{-}\theta_{\floor{\bldv}_{d,\bldx}})\Big)\\
\phantom{=}&+\smashoperator{\sum_{d,\bldv}}\Phi_\bldv(\bldx)(\theta_{\ceil{\bldv}_{d,\bldx}}{-}\theta_{\floor{\bldv}_{d,\bldx}})\floor{\bldx[d]}.
\end{align}
Now, using~(\ref{eqn:phi-def}) and defining $\lambda(v,x){=}1{+}(x{-}v)(-1)^{I_{v=\floor{x}}}$ for $x{\in}\RR$ and $v{\in}\NN$, we have for each $\bldv\in\N(\bldx)$, with $\bldv{=}\ceil{\bldv}_{d,\bldx}$ 
\begin{align*}
&\Big(\Phi_{\bldv}(\bldx)+\Phi_{\floor{\bldv}_{d,\bldx}}(\bldx)\Big)(\bldx[d]-\floor{\bldx[d]})\\
&\quad = (\bldx[d]-\floor{\bldx[d]})
\smashoperator[l]{\sum_{\bldw\in\{\bldv,\floor{\bldv}_{d,\bldx}\}}}\prod_{i=1}^{D}\lambda(\bldw[i],\bldx[i]) \\
&\quad = (\bldx[d]-\floor{\bldx[d]})\Big(\prod_{i\neq d}\lambda(\bldv[i],\bldx[i])\Big)\cdot\\
&\quad \phantom{=} \quad\Big(\lambda({\floor{\bldx[d]}{+}1},\bldx[d])+
          \lambda({\floor{\bldx[d]}},\bldx[d])\Big),
\end{align*}
where to get the last equality, observe that for $i\neq d$, the $i$th entry of $\bldv$ and $\floor{\bldv}_{d,\bldx}$ is the same.
Noting that $\lambda(\floor{\bldx[d]}{+}1,\bldx[d])+\lambda(\floor{\bldx[d]},\bldx[d]) = 1$ and that $\bldx[d]-\floor{\bldx[d]}=\lambda(\bldx[d],\bldv[d])$, we get
\begin{align}\label{eqn:unimodality-sum-of-phis}
\nonumber \Big(\Phi_{\bldv}(\bldx)+\Phi_{\floor{\bldv}_{d,\bldx}}(\bldx)\Big)(\bldx[d]-\floor{\bldx[d]}) &= \prod_{i=1}^{D}\lambda(\bldv[i],\bldx[i]) \\
&= \Phi_\bldv(\bldx)
\end{align}

Plugging~(\ref{eqn:unimodality-sum-of-phis}) 
into~(\ref{eqn:unimodality-f'}), we get
\begin{align*}
\nonumber f'_\bldx(1) 
& = \smashoperator{\sum_{d,\bldv:\bldv=\ceil{\bldv}_{d,\bldx}}}
\Phi_{\bldv}(\bldx)(\theta_{\ceil{\bldv}_{d,\bldx}}{-}\theta_{\floor{\bldv}_{d,\bldx}}) \\
&\phantom{=}+\sum_{d,\bldv}\Phi_\bldv(\bldx)(\theta_{\ceil{\bldv}_{d,\bldx}}{-}\theta_{\floor{\bldv}_{d,\bldx}})\floor{\bldx[d]}
\\ \nonumber 
& = \sum_{d,\bldv}\Big(\Phi_\bldv(\bldx)(\theta_{\ceil{\bldv}_{d,\bldx}}{-}\theta_{\floor{\bldv}_{d,\bldx}}) \cdot(I_{\bldv=\ceil{\bldv}_{d,\bldx}}{+}\floor{\bldx[d]}) \Big)
\\
& = \smashoperator{\sum_{\bldv\in\N(\bldx)}}\Phi_\bldv(\bldx)\sum_{d=1}^{s}(\theta_{\ceil{\bldv}_{d,\bldx}}{-}\theta_{\floor{\bldv}_{d,\bldx}})\bldv[d],
\end{align*}
where the last equality holds since, for each $\bldv{\in}\N(\bldx)$, $I_{\bldv{=}\ceil{\bldv}_{d,\bldx}}{+}\floor{\bldx[d]}{=}\bldv[d]$.

Hence $f'_\bldx(1)$ is a multi-linear interpolation of the values $\{\sum_{d=1}^{s}(\theta_{\ceil{\bldv}_{d,\bldx}}{-}\theta_{\floor{\bldv}_{d,\bldx}})\bldv[d]\}_{\bldv}$ on $\N(\bldx)$. Thus requiring that it would be nonpositive for all $\bldx\in\RR^D$ is equivalent to requiring that
$$
\sum_{d=1}^{s}(\theta_{\ceil{\bldv}_{d,\bldx}}{-}\theta_{\floor{\bldv}_{d,\bldx}})\bldv[d] \leq 0,\;\forall \N(\bldx),\bldv{\in}\N(\bldx).
$$
It's easy to verify that these are precisely the inequalities in Statement~2. 
\end{proof}

\subsection{Unimodal Lattice Not Sufficient For a GON To Be Unimodal}

\PropGONIsNotUnimodal

\begin{proof} 
Our proof is by counterexample. Let $f$ be the function of the 2D lattice with size $(3,3)$ and vertex values: $\theta_{(0,0)}=3,\theta_{(-1,0)}=\theta_{(1,0)}=2,\theta_{(0,-1)}=\theta_{(0,1)}=0,\theta_{(-1,-1)}=\theta_{(1,-1)}=\theta_{(-1,1)}=\theta_{(1,1)}=1$. It's easy to verify that equation~(\ref{eqn:lattice-unimodality-cond}) of Lemma~1 holds for s=2. Thus $f$ satisfies the unimodality shape constraint with maximizer $(0,0)$. Now, let
$c_1:[0,3]\rightarrow[-1,1]$ and $c_2:[0,3]\rightarrow[-1,1]$ be the PLFs given by:
\begin{equation*}
    c_1(x) = \left\{ 
    \begin{array}{ll}
    x-1 & \mbox{if $0\leq x < 1$} \\
    (x-1)/2 & \mbox{if $1\leq x \leq 3$} \\
    \end{array}
    \right.,
\end{equation*}
and
\begin{equation*}
    c_2(x) = \left\{ 
    \begin{array}{ll}
    x-1 & \mbox{if $0\leq x < 2$} \\
    1 & \mbox{if $2\leq x \leq 3$} \\
    \end{array}
    \right..
\end{equation*}
Let $g(x,y)=f(c_1(x),c_2(y))$. Then it can be easily verified that the global maximizer of $g$ is at $(1,1)$ and it is unique. Thus for $g$ to satisfy the unimodal shape constraint, it must do so with maximizer $(1,1)$. However $g$ is not decreasing along the ray $r(t)=(1,1)+t(1,1)$, since $g(r(1))=g(2,2)=f(1/2,1)=(\theta_{(1,1)}+\theta_{(0,1)})/2=1/2$ and $g(r(2))=g(3,3)=f(1,1)=\theta_{(1,1)}=1$.

See Figure~\ref{fig:gonisnotunimodal} for the illustration of the $f$ and $g$ functions.

\begin{figure}[!t]
\centering
\begin{tabular}{c}
\includegraphics[width=5in]{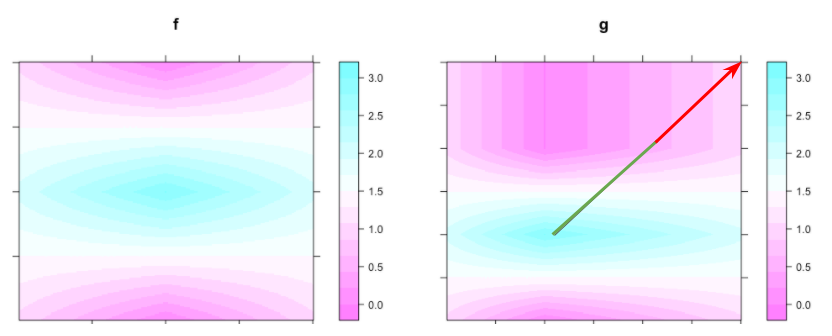}
\end{tabular}
\caption{Illustrated counterexample for Prop. 4 with two-dimensional functions $f$ and $g$ shown over the domain $[-1,1]^2$. The function $f$ is unimodal, but the resulting GON function $g$ is not, for example the shown ray starts at the global minimum but the function along that ray is  only monotonically decreasing for the green part, the function is decreasing along the red part.}
\label{fig:gonisnotunimodal} 
\end{figure}

\end{proof}

\subsection{Ensemble of Unimodal Functions is Unimodal}
\PropEnsembleOfUnimodalsIsUnimodal*
\begin{proof}
Let $\bldz(r)=r\bldv$, $r\geq0$ be a ray in $I^D$ originating in $\bldzero$ for some $\bldv\in\RR^{D}$. We need to show that $u(\bldz(r))=a_0 + \sum_t{a_t u_t(\pi_t(\bldz(r)))}$ is decreasing for $r\geq0$. Since $\pi_t(\bldz(r))=r\pi_t(\bldv)$, $r\geq0$ is a ray in $\RR^{Q}$ in direction $\pi_t(\bldv)$ originating in $\bldzero\subseteq\calS_Q$, it follows by the unimodality of each $u_t$ that $u_t(\pi_t(\bldz(r)))$ is decreasing for $r\geq0$. The result now follows from the fact that a conical sum of decreasing functions is decreasing.
\end{proof}

\section{Details for Monarchs' Reigns Experiments}
We provide more details on the data and experimental results.  

\subsection{Data Details for Monarchs' Reigns Experiments}
The data can be downloaded at www.kaggle.com/senzhaogoogle/kingsreign. 

All fifty dynasties were sampled from across the globe and from ancient to modern times. The original Monarchs' Reigns dataset \cite{Feldman:2014} (also known as Kings' Reigns, but some of the monarchs were queens or had other titles) consists of 30 royal dynasties, for example the 36 monarchs of the Ottoman Empire from 1299-1922, the 15 monarchs of the Kings of Larsa from 1961 BC to 1674 BC, and the 4 monarchs of the Zulu Dynasty of 1816-1879. We added a test set of 20 additional royal dynasties using the same methodology used for the original dataset based on conversations with the original dataset creator Kyle Stewart (based on conversations about methodology).  Example test set dynasties are the 5 monarchs of the 18th century Hotak dynasty in Afghanistan, and the 27 monarch Joseon dynasty of Korea that ended in 1910.  All information came from Wikipedia. We will provide a Kaggle notebook for the complete train and test datasets. 

The train and test datasets have the following known sampling biases:
\begin{itemize}
    \item Dynasties for which there were more complete and well-organized records on Wikipedia were more likely to be sampled. This likely caused under-sampling of pre-Columbian American dynasties, for example. 
    \item An effort was made to sample geographically diverse dynasties, which may have caused under-sampling of some regions and over-sampling of other regions with regards to population.
    \item An effort was made to sample dynasties across time, which may have caused under-sampling of some timeframes and over-sampling of others with regards to population.
    \item Current dynasties where the last monarch is still reigning were not sampled.
\end{itemize}

We note that our use of this data simplifies a number of potentially important factors about the stability of dynasties. For example, in monogamous cultures, it was more difficult to ensure a direct heir than in polygamous cultures \cite{Dynasties:2015}.  A second issue is simply the definition of dynastic boundaries: what counts as a new dynasty, and has that criteria been sufficiently uniformly applied to the diverse dynasties in this dataset?  A third issue is we treated the dynasties as though they were samples drawn IID from the same distribution, but the general reduction in violence over documented history \cite{BetterAngels} might imply a shifting distribution towards more stable dynasties, given that many change-overs were due to violence.

\subsection{Experimental Details for Monarchs' Reigns}

The train set had $N=30$ dynasties, and the test set had $20$ dynasties. 
For each method, we cross-validated over 18 choices of hyperparameters by leave-one-out cross-validation: we left out one-dynasty at a time and trained a model with each choice of hyperparameters on the other 29 dynasties. For each trained model and left-out dynasty, the predicted maximizer was computed as: $\hat{x} = \arg \max_{x \in \mathcal{X}_{\textrm{Left-out}}} h(x)$, and we scored $\hat{x}$ by the actual number of years reigned by that monarch in the left-out dynasty. Averaging those scores over the 30 rounds of one-dynasty-left-out formed the overall validation score for that hyperparameter choice.  Tables in the Appendix list the 18 hyperparameter choices and corresponding validation scores for each method.

The test metric is the same as the cross-validation metric: for each trained model and each test dynasty, the predicted maximizer was computed as: $\hat{x} = \arg \max_{x\in\mathcal{X}_{\textrm{Test}}} h(x)$, and we scored $\hat{x}$ by the actual number of years reigned by that monarch in that test dynasty. Averaging those scores over the 20 test dynasties formed the overall test score for that method. 

Table \ref{tab:kings} shows that the GON model achieved the best test score, followed by the GPR. Note that while both DNN and GON predict a 6th monarch will rule longest, their test scores differ because they made different predictions for the maximizer for test dynasties that have fewer than 6 monarchs, as can be seen in Figure \ref{fig:kings}.  

\begin{table}[!t]
\caption{Longest Reign: Models With Best Validation Scores. Units are years. Bold is best.}
\label{tab:kings}
\centering
\begin{tabular}{llllll}
\toprule   
Model     & Train Set: &  Test Set: Mean Actual   & Global  \\
 & Root MSE  &   Reign of Model's Arg Max's & Arg Max\\
\midrule
DNN  & 14.67 & 15.05 & 6th monarch \\
FICNN     &14.89 & 14.75 &  1st monarch\\
GPR  & 15.54 & 16.40 & 7th monarch \\
GON   &14.80 & \textbf{16.95}  &  6th monarch\\
\bottomrule
\end{tabular}
\end{table}

\subsection{Cross-Validation Scores For Different Hyperparameters}\label{sec:kingscv}
The complete cross-validation scores are shown for all the tried hyperparameters in Tables \ref{tab:kingsReignsGON}, \ref{tab:kingsReignsICNN} and \ref{tab:kingsReignsDNN}. 

\begin{table*}[!t]
\caption{Monarchs' Reigns: GON Validation Scores Over Hyperparameters. Bold is the highest validation score for this model type.}
\label{tab:kingsReignsGON}
\centering
\begin{tabular}{llll}
\toprule                \\
Model   & Number Keypoints in $f$ & Number Keypoints in $c$ &  Validation Score\\
\midrule
GON & 3 & 2 & 21.23 \\
GON & 3 & 3 & 20.97 \\
GON & 3 & 5 & 25.30 \\
GON & 3 & 7 &  25.26 \\
GON & 3 & 9 & \textbf{26.00} \\
GON & 3 & 11 & 23.8 \\
GON & 5 & 2 & 22.97 \\
GON & 5 & 3 & 18.33 \\
GON & 5 & 5 & 25.27 \\
GON & 5 & 7 &  20.06 \\
GON & 5 & 9 &  22.80 \\
GON & 5 & 11 & 20.93 \\
GON & 7 & 2 & 22.47 \\
GON & 7 & 3 & 22.46 \\
GON & 7 & 5 & 18.33 \\
GON & 7 & 7 & 18.50 \\
GON & 7 & 9 & 20.77 \\
GON & 7 & 11 & 18.73 \\
GON & 9 & 2 & 22.46 \\
GON & 9 & 3 & 18.30 \\
GON & 9 & 5 & 19.73 \\
GON & 9 & 7 & 19.20  \\
GON & 9 & 9 & 21.13 \\
GON & 9 & 11 & 22.10 \\
\midrule
\bottomrule
\end{tabular}
\end{table*}

\begin{table*}[!t]
\caption{Monarchs' Reigns: FICNN Validation Scores Over Hyperparameters. Bold is the highest validation score for this model type.}
\label{tab:kingsReignsICNN}
\centering
\begin{tabular}{llll}
\toprule                \\
Model   & Number Layers & Number Hidden Nodes & Validation Score\\
\midrule
FICNN  & 3 & 2 & 22.47 \\
FICNN  & 3 & 4 &  22.47 \\
FICNN  & 3 & 8 & 22.20 \\
FICNN  & 3 & 16 & 22.00 \\
FICNN  & 3 & 32 & 21.20 \\
FICNN  & 3 & 64 &  20.70 \\
FICNN  & 4 & 2 & 22.93 \\
FICNN  & 4 & 4 &  \textbf{23.17} \\
FICNN  & 4 & 8 & 21.83 \\
FICNN  & 4 & 16 & 21.63 \\
FICNN  & 4 & 32 & 21.20 \\
FICNN  & 4 & 64 &  20.50\\
FICNN  & 5 & 2 & 20.36 \\
FICNN  & 5 & 4 &  22.30 \\
FICNN  & 5 & 8 & 20.70 \\
FICNN  & 5 & 16 & 20.96 \\
FICNN  & 5 & 32 & 19.40 \\
FICNN  & 5 & 64 &  20.70\\
FICNN  & 6 & 2 & 22.06 \\
FICNN  & 6 & 4 &  20.77 \\
FICNN  & 6 & 8 & 21.83 \\
FICNN  & 6 & 16 & 19.73 \\
FICNN  & 6 & 32 & 21.50 \\
FICNN  & 6 & 64 &  19.73\\
\bottomrule
\end{tabular}
\end{table*}

\begin{table*}[!t]
\caption{Monarchs' Reigns: DNN Validation Scores Over Hyperparameters. Bold is the highest validation score for this model type.}
\label{tab:kingsReignsDNN}
\centering
\begin{tabular}{llll}
\toprule                \\
Model   & Number Layers & Number Hidden Nodes & Validation Score\\
\midrule
DNN  & 3 & 2 & 21.87 \\
DNN  & 3 & 4 & 18.03 \\
DNN  & 3 & 8 &  17.53 \\
DNN  & 3 & 16 & 19.73 \\
DNN  & 3 & 32 & 21.26  \\
DNN  & 3 & 64 &  22.4 \\
DNN  & 4 & 2 &  19.97 \\
DNN  & 4 & 4 &  20.26  \\
DNN  & 4 & 8 &  20.43 \\
DNN  & 4 & 16 &  19.4  \\
DNN  & 4 & 32 &  20.4 \\
DNN  & 4 & 64 &  21.20\\
DNN  & 5 & 2 & 22.46 \\
DNN  & 5 & 4 &  18.76 \\
DNN  & 5 & 8 & 22.16  \\
DNN  & 5 & 16 & 21.03 \\
DNN  & 5 & 32 &  22.47\\
DNN  & 5 & 64 & 24.13 \\
DNN  & 6 & 2 &  22.46 \\
DNN  & 6 & 4 & 19.76  \\
DNN  & 6 & 8 &  21.80  \\
DNN  & 6 & 16 &  24.13 \\
DNN  & 6 & 32 &  \textbf{24.70} \\
DNN  & 6 & 64 &  24.20 \\
\midrule
\bottomrule
\end{tabular}
\end{table*}

\begin{table*}[!t]
\caption{Monarchs' Reigns: GPR Validation Scores Over Hyperparameters. Bold is the highest validation score for this model type.}
\label{tab:kingsReignsGPR}
\centering
\begin{tabular}{llll}
\toprule                \\
Model   & $\alpha$ &  Validation Score\\
\midrule
GPR & $\alpha = 1e-12$ & 18.97 \\
GPR & $\alpha = 1e-11$ & 17.83 \\
GPR & $\alpha = 1e-10$ & 17.46 \\
GPR & $\alpha = 1e-9$ & 16.83 \\
GPR & $\alpha = 1e-8$ & 18.13 \\
GPR & $\alpha = 1e-7$ & 20.7 \\
GPR & $\alpha = 1e-6$ & 21.43 \\
GPR & $\alpha = 1e-5$ & 21.6 \\
GPR & $\alpha = 1e-4$ & 14.93 \\
GPR & $\alpha = 1e-3$ & 18.97 \\
GPR & $\alpha = 1e-2$ & 21.7 \\
GPR & $\alpha = 1e-1$ & 22.47\\
GPR & $\alpha = 1$ & 22.47 \\
GPR & $\alpha = 10$ & 22.47 \\
GPR & $\alpha = 100$ & \textbf{23.27} \\
GPR & $\alpha = 1e3$ & 19.63 \\
GPR & $\alpha = 1e4$ & 19.70 \\
GPR & $\alpha = 1e5$ & 19.70 \\
GPR & $\alpha = 1e6$ & 19.70 \\
GPR & $\alpha = 1e7$ & 19.70 \\
GPR & $\alpha = 1e8$ & 19.70 \\
GPR & $\alpha = 1e9$ & 19.70 \\
GPR & $\alpha = 1e10$ & 19.70 \\
GPR & $\alpha = 1e11$ & 19.70 \\
\midrule
\bottomrule
\end{tabular}
\end{table*}

\section{Details for Puzzles Experiments}
The hyperparameter choices were designed to give a range of flexibility. For the FICNN and DNN models, choices were either $3$ or $4$ layers ($3$ layers being the default in \cite{Zico:2017}), and either $\{2,4,6,8\}$ hidden nodes. The GPR hyperparameter was the sklearn standard covariance matrix additive smoothing parameter $\alpha$, ranging from $1e-6$ to 10 in steps of 10.  All GON models used a unimodal $3 \times 3$ lattice layer for $f(x)$, and varied the number of keypoints in $c(x)$'s PLFs from $K=2$ to $K=9$. Because the first and last PLF keypoint are fixed to map to the lattice layer's input domain, the $K=2$ case is equivalent to not having a first layer.  Any ties were decided in favor of the hyperparameters corresponding to a more-regularized model.

Table \ref{tab:puzzlesHalf} and \ref{tab:puzzlesOtherHalf} reports actual sales of the highest-predicted validation and test puzzles. The GON was most accurate in predicting the best-selling test puzzle, followed by the DNN and GON. The GPR model chose a test puzzle that was actually a terrible seller. 

Our test metric was limited to the test set of puzzles for which there was 2019 sales numbers. In practice though, the business would like to use such a model for guidance as to which new puzzles they should create. For such use, we should ask if the global maximizer is reasonable. As seen in Figure \ref{fig:bestPuzzles}, the FICNN and DNN models extrapolated poorly from a popular small puzzle in the train set, leading those models to predict that the global optimizer would be a jigsaw puzzle with zero pieces, which is not reasonable guidance. We questioned whether this was simply bad luck in selecting the hyperparameters, but in fact, 5 of the 8 FICNN models trained predicted the argmax at 0 pieces (see Table \ref{tab:puzzlesHalf} in the Appendix). The DNN also only gave reasonable answers for the global maximizer for 3 of its 8 hyperparameter choices. 

The five most-flexible GON models consistently predicted a global optimizer would be a puzzle with 190-230 pieces and artwork from around the year 2000. Partners at Artifact said that based on their ten years of sales experience, such puzzles do tend to sell best. 

We also note the GON models also generally predicted the best year for art was 2000, which is at the edge of the input domain, which confirms the proposed unimodal shape constraints do not block fitting models with their maximizer on the edge of the input domain.

\begin{table*}[!t]
\caption{New Puzzle Sales: Results for Different Hyperparameters for DNN and FICNN. As marked, the DNN model sometimes came out ``flat", that is, it predicted the same sales for all inputs. Ties broken in favor of the smaller/smoother model.}
\label{tab:puzzlesHalf}
\centering
\begin{tabular}{llllll}
\toprule                \\
Model      & Actual Sales of  & Actual Sales of  & Global Arg Max\\
 &  Highest-Scored   &  Highest-Scored &   \\
  & Validation Puzzle & Test Puzzle &  \\
\midrule
DNN 3 layers, 2 hid.  & flat model & -- & -- \\
DNN 3 layers, 4 hid.  & 21 & 30 & 1200 pieces, year 2000 \\
DNN 3 layers, 6 hid.   & 74 & 182 & 192 pieces, year 2000 \\
DNN 3 layers, 8 hid.   & \textbf{88} & 173 & 0 pieces, year 2000 \\
DNN 4 layers, 2 hid.  & flat model & -- & --  \\
DNN 4 layers, 4 hid.  & 0 & 7 & 0 pieces, year 1500 \\
DNN 4 layers, 6 hid.  & 10 & 30 & 192 pieces, year 2000  \\
DNN 4 layers, 8 hid.  & 16 & 164 & 353 pieces, year 2000 \\
\midrule
FICNN 3 layers, 2 hid.   & 43 & 173 & 0 pieces, year 2000 \\
FICNN 3 layers, 4 hid.  & \textbf{88} & 173 & 0 pieces, year 2000 \\
FICNN 3 layers, 6 hid.   & 88 & 173 & 0 pieces, year 2000 \\
FICNN 3 layers, 8 hid.   & 74 & 182 & 192 pieces, year 2000 \\
FICNN 4 layers, 2 hid.   & 88 & 173 & 0 pieces, year 2000 \\
FICNN 4 layers, 4 hid.  & 74 & 13 & 0 pieces, year 2000 \\
FICNN 4 layers, 6 hid.  & 74 & 182 & 0 pieces, year 2000  \\
FICNN 4 layers, 8 hid.  & 88 & 173 & 0 pieces, year 2000 \\
\bottomrule
\end{tabular}
\end{table*}

\begin{table*}[!t]
\caption{New Puzzle Sales: Results for Different Hyperparameters for GPR and GON. Ties broken in favor of the smaller/smoother model.}
\label{tab:puzzlesOtherHalf}
\centering
\begin{tabular}{llllll}
\toprule                \\
Model      & Actual Sales of  & Actual Sales of  & Global Arg Max\\
 &  Highest-Scored   &  Highest-Scored &   \\
  & Validation Puzzle & Test Puzzle & \\
\midrule
GPR $\alpha = 1e-6$  & 21 & 3 & 168 pieces, year 1500\\ 
GPR $\alpha = 1e-5$ & 21 & 182 & 242 pieces, year 2000 \\
GPR $\alpha = 1e-4$ & 74  & 182  & 242 pieces, year 2000 \\
GPR $\alpha = 1e-3$  & 74 & 182  &  242 pieces, year 2000 \\
GPR $\alpha = 1e-2$  & 88  & 173 & 68 pieces, year 2000\\
GPR $\alpha = 1e-1$  & 43  & 173  & 68 pieces, year 2000\\
GPR $\alpha = 1$ & 43  & 173  & 68 pieces, year 2000\\
GPR $\alpha = 10$  & \textbf{88}  & 2  & 146 pieces, year 2000 \\
\midrule
GON  2kp     & 43 & 1 & 600 pieces, year 1700\\
GON  3kp  & 31 & 21 & 502 pieces, year 1400\\
GON  4kp  & \textbf{76} & 182 & 230 pieces, year 2000\\
GON  5kp  & 74 & 182 & 190 pieces, year 2000\\
GON  6kp  & 13 & 182 & 212 pieces, year 2000\\
GON  7kp   & 74 & 182 & 191 pieces, year 2000\\
GON  8kp   & 74 & 182 & 194 pieces, year 2000\\
GON  9kp   & 74 & 182 & 213 pieces, year 2000\\
\bottomrule
\end{tabular}
\end{table*}

\section{Details for Wine Experiments}

Tables~\ref{tab:wineDNN}-\ref{tab:wineGON} summarize the validation scores of DNN, FICNN and GON over hyperparameters.

\begin{figure*}[!t]
\centering
\begin{tabular}{c}
\includegraphics[width=5.0in]{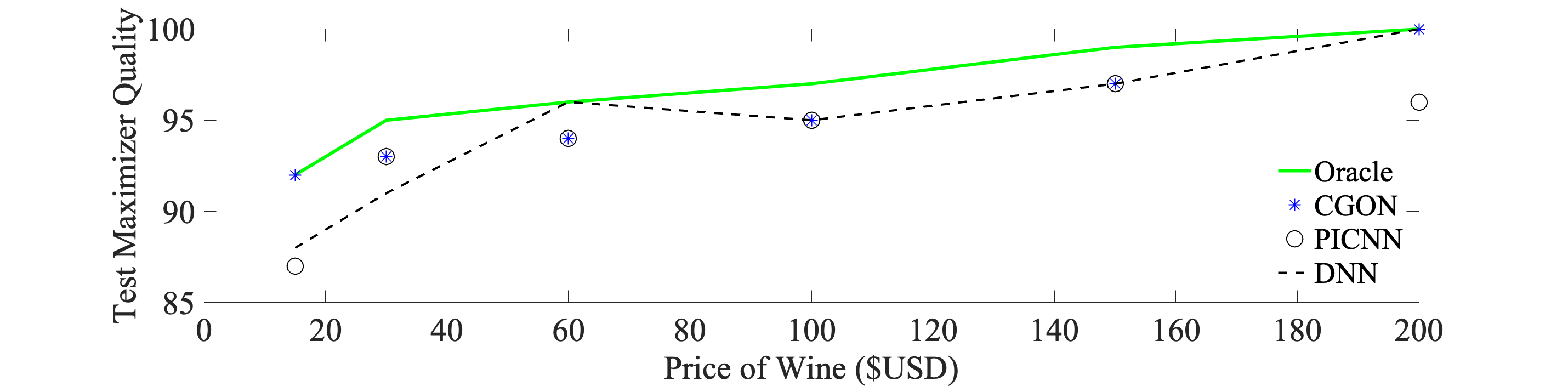} 
\end{tabular}
\caption{Quality of the predicted highest-quality wine conditioned on price. The oracle marks the true best wine from the test set for each price point.}
\label{fig:wineByPrice} 
\end{figure*}

Figure \ref{fig:wineByPrice} shows the results for the experiments conditioned on price. 

\begin{table*}[!t]
\caption{Best Wine: DNN Validation Scores Over Hyperparameters. Bold is the highest validation score for this model type, with ties broken in favor of the smallest model with that validation score. Surprisingly, the DNN consistently chose the same poor test wine as its predicted best. Analysis showed that the DNN's were extrapolating poorly in the high price part of the feature space and putting too much faith in high price as a signal of quality, and that the DNN's prediction is the most expensive test wine.}
\label{tab:wineDNN}
\centering
\begin{tabular}{lllll}
\toprule                \\
Model & Val  & Train  & Test  & Test Maximizer \\
 & Score & MSE & Score & \\
\midrule
DNN: 2 layers, 2 nodes  & \textbf{97} & 2.54 & 88 & \$3300, acid, juicy, tannin, France \\
DNN: 2 layers, 4 nodes  & 97 & 2.53 & 88 &  \$3300, acid, juicy, tannin, France \\
DNN: 2 layers, 8 nodes  & 97 & 2.54  & 88 & \$3300, acid, juicy, tannin, France \\
DNN: 2 layers, 16 nodes  & 97 &  2.47 & 88 &  \$3300, acid, juicy, tannin, France\\
DNN: 2 layers, 32 nodes  & 97 & 2.30  & 88 & \$3300, acid, juicy, tannin, France \\
DNN: 2 layers, 64 nodes  & 92  & 2.24 & 94 & \$1100, complex, earth, lees, tight, Austria \\
DNN: 3 layers, 2 nodes  & 97 & 2.55 & 88 & \$3300, acid, juicy, tannin, France \\
DNN: 3 layers, 4 nodes  & 97 & 2.55 & 88 &  \$3300, acid, juicy, tannin, France \\
DNN: 3 layers, 8 nodes  & 97 &  2.55 & 88 & \$3300, acid, juicy, tannin, France \\
DNN: 3 layers, 16 nodes  & 97 &  2.27 & 88 &  \$3300, acid, juicy, tannin, France\\
DNN: 3 layers, 32 nodes  & 97 &  2.23 & 88 & \$3300, acid, juicy, tannin, France \\
DNN: 3 layers, 64 nodes  & 97  & 2.24 & 88 & \$3300, acid, juicy, tannin, France \\
DNN: 4 layers, 2 nodes  & 97 & 2.54 & 88 & \$3300, acid, juicy, tannin, France \\
DNN: 4 layers, 4 nodes  & 97 & 2.53 & 88 &  \$3300, acid, juicy, tannin, France \\
DNN: 4 layers, 8 nodes  & 97 & 2.27 & 88 & \$3300, acid, juicy, tannin, France \\
DNN: 4 layers, 16 nodes  & 97 & 2.23 & 88 &  \$3300, acid, juicy, tannin, France\\
DNN: 4 layers, 32 nodes  & 97 & 2.23 & 88 & \$3300, acid, juicy, tannin, France \\
DNN: 4 layers, 64 nodes  & 97  & 2.17 & 88 & \$3300, acid, juicy, tannin, France \\
\midrule
\bottomrule
\end{tabular}
\end{table*}

\begin{table*}[!t]
\caption{Best Wine: FICNN Validation Scores Over Hyperparameters. Bold is the highest validation score for this model type, with ties broken in favor of the smallest model with that validation score. Like the DNN, analysis showed the FICNN tended to overfit high price as a sign of quality and often chose the most expensive test wine, which actually did not have high points.}
\label{tab:wineICNN}
\centering
\begin{tabular}{lllll}
\toprule                \\
Model & Val  & Train  & Test & Test Maximizer \\
 & Score & MSE & Score & \\
\midrule
FICNN: 2 layers, 2 nodes  & 97 & 2.52 & 88 & \$3300, acid, juicy, tannin, France \\
FICNN: 2 layers, 4 nodes  & 97 & 2.29 & 88 & \$3300, acid, juicy, tannin, France \\
FICNN: 2 layers, 8 nodes  & 91 & 2.33  & 95 & \$412, jam, opulent, France \\
FICNN: 2 layers, 16 nodes  & 92 &  2.27 & 100 & \$848, acid, hint of, opulent, toast, France  \\
FICNN: 2 layers, 32 nodes  & \textbf{98} & 2.20  & 94 & \$1100, complex, earth, lees, tight, Austria \\
FICNN: 2 layers, 64 nodes  & 96  & 2.19 & 94 &  \$1100, complex, earth, lees, tight, Austria\\
FICNN: 3 layers, 2 nodes  & 97 & 2.53 & 88 & \$3300, acid, juicy, tannin, France \\
FICNN: 3 layers, 4 nodes  & 96 & 2.30 & 94 &  \$1100, complex, earth, lees, tight, Austria  \\
FICNN: 3 layers, 8 nodes  & 96 &  2.25 & 94 & \$1100, complex, earth, lees, tight, Austria \\
FICNN: 3 layers, 16 nodes  & 96 &  2.23 & 94 & \$1100, complex, earth, lees, tight, Austria \\
FICNN: 3 layers, 32 nodes  & 91 &  2.24 & 96 & \$351  oak, tannin, tight, toast, Spain \\
FICNN: 3 layers, 64 nodes  & 92  & 2.30 & 94 & \$1100, complex, earth, lees, tight, Austria \\
FICNN: 4 layers, 2 nodes  & 97 & 2.40 & 94 & \$1100, complex, earth, lees, tight, Austria \\
FICNN: 4 layers, 4 nodes  & 97 & 2.24 & 94 & \$900, elegant, Italy \\
FICNN: 4 layers, 8 nodes  & 97 & 2.26 & 85 & \$320, acid, crisp, Romania \\
FICNN: 4 layers, 16 nodes  & 97 & 2.28 & 88 &  \$3300, acid, juicy, tannin, France\\
FICNN: 4 layers, 32 nodes  & 97 & 2.47 & 88 & \$3300, acid, juicy, tannin, France \\
FICNN: 4 layers, 64 nodes  &  97 & 2.47 & 88 & \$3300, acid, juicy, tannin, France \\
\midrule
\bottomrule
\end{tabular}
\end{table*}


\begin{table*}[!t]
\caption{Best Wine: GON Validation Scores Over Hyperparameters. Bold is the highest validation score for this model type, with ties broken in favor of the smallest model with that validation score.}
\label{tab:wineGON}
\centering
\begin{tabular}{lllll}
\toprule                \\
Model & Val  & Train  & Test  & Test Maximizer \\
 & Score & MSE & Score & \\
\midrule
GON 100 2D lattices, 5kp & 92 & 2.31 & 95 & \$100 acid, cassis, complex, refined\\
 &&&& structure, tannin, velvet, US\\
GON  100 2D lattices, 9kp & 93 & 2.29 & 97 & \$375, acid, bright, complex, elegant, \\ 
&&&& refined, structure, tannin, Italy\\
GON  100 2D lattices, 13kp  & 93 & 2.31 & 97 & \$375, acid, bright, complex, elegant \\ 
&&&& refined, structure, tannin, Italy\\
GON  200 2D lattices, 5kp  & 97 & 2.30 & 97 & \$165 acid, cassis, complex, mineral\\ 
&&&& oak, refined, structure, tannin, US\\
GON  200 2D lattices, 9kp  & \textbf{98} & 2.28 & 97 & \$375, acid, bright, complex, elegant \\ 
&&&& refined, structure, tannin, Italy\\
GON  200 2D lattices, 13kp  & 98 & 2.26 & 97 & \$375, acid, bright, complex, elegant \\ 
&&&& refined, structure, tannin, Italy\\
GON  400 2D lattices, 5kp  & 96 & 2.32 & 95 & \$100, acid, cassis, complex\\ 
&&&& refined, structure, tannin, velvet, US\\

GON  400 2D lattices, 9kp  & 97  &  2.27 & 97 &  \$375, acid, bright, complex, elegant \\ 
&&&& refined, structure, tannin, Italy\\

GON 400 2D lattices, 13kp  & 97 & 2.25 & 94 &  \$1100, complex, earth, lees \\ 
&&&& tight, Austria \\ 

GON  800 2D lattices, 5kp  & 97 & 2.28  & 97 &\$375, acid, bright, complex, elegant \\ 
&&&& refined, structure, tannin, Italy \\ 

GON  800 2D lattices, 9kp  & 98 & 2.26 & 94 & \$1100, complex, earth, lees \\ 
&&&& tight, Austria \\

GON  800 2D lattices, 13kp  & 93 & 2.24 & 97 &\$375, acid, bright, complex, elegant \\ 
&&&& refined, structure, tannin, Italy \\ 

GON  1600 2D lattices, 5kp  &97 &  2.26 &  96 & \$180 butter, complex, lees, \\
&&&& mineral US\\ 
GON  1600 2D lattices, 9kp  & 97 & 2.37 & 94 & \$1100, complex, earth, lees \\ 
&&&& tight, Austria \\ 

GON  1600 2D lattices, 13kp  & 94 & 2.22 & 96 & \$450, cream, dense, mineral \\
&&&& tight, France \\
\bottomrule
\end{tabular}
\end{table*}

\section{Details for Hyperparameter Optimization for Image Classifiers}
As mentioned in the main paper, image classifers shown in the main paper are trained for $e\in[1,20]$ epochs. We use ADAM with the default learning rate of 0.001 with a batch size of 128 to train the classifiers.

For GON and CGON, we use an ensemble of $D$ unimodal lattices. All methods are trained for 250 epochs.

Hyperparameters for the optimizers are validated based on 5-fold MSE, which are summarized in Table \ref{tab:hparamshparams} below.

\begin{table*}[!t]
\caption{Hyperparameters of optimizers.}
\label{tab:hparamshparams}
\centering
\begin{tabular}{lrrrrrrrr}
\toprule 
& \multicolumn{4}{c}{Global Optimization} &  \multicolumn{4}{c}{Conditional Global Optimization} \\
\cmidrule(lr){2-5} \cmidrule(lr){6-9}
Rosenbrock &  GON & FICNN & DNN & GPR &  CGON & PICNN & DNN & GPR \\
\midrule
PLF kps per input $K$ & 5 & - & - & - & 5 & - & - & - \\
Lattice kps per input $V$ & 3 & - & - & - & 3 & - & - & - \\
Inputs each lattice fuses & 2 & - & - & - & 2 & - & - & - \\
Num hidden layers & - & 1 & 1 & - & - & 1 & 1 & - \\
Num hidden nodes & - & 256 & 32 & - & - & 32 & 32 & - \\
$\alpha$ in GPR  & - & - & - & 0.01 & - & - & - & 0.01 \\
\bottomrule
\end{tabular}
\end{table*}

\section{Details for Simulations with Standard Global Optimization Functions}
We ran simulations on two standard benchmark functions, the banana-shaped Rosenbrock function, and the pocked-convex Griewank function, to compare GON against FICNN, DNN, GPR and sample best. For conditional global optimization problems, we compared CGON against PICNN, DNN and GPR.

\begin{figure}[!t]
\centering
\begin{tabular}{cc}
\includegraphics[trim={0 0 0 1cm},clip,width=1.6in]{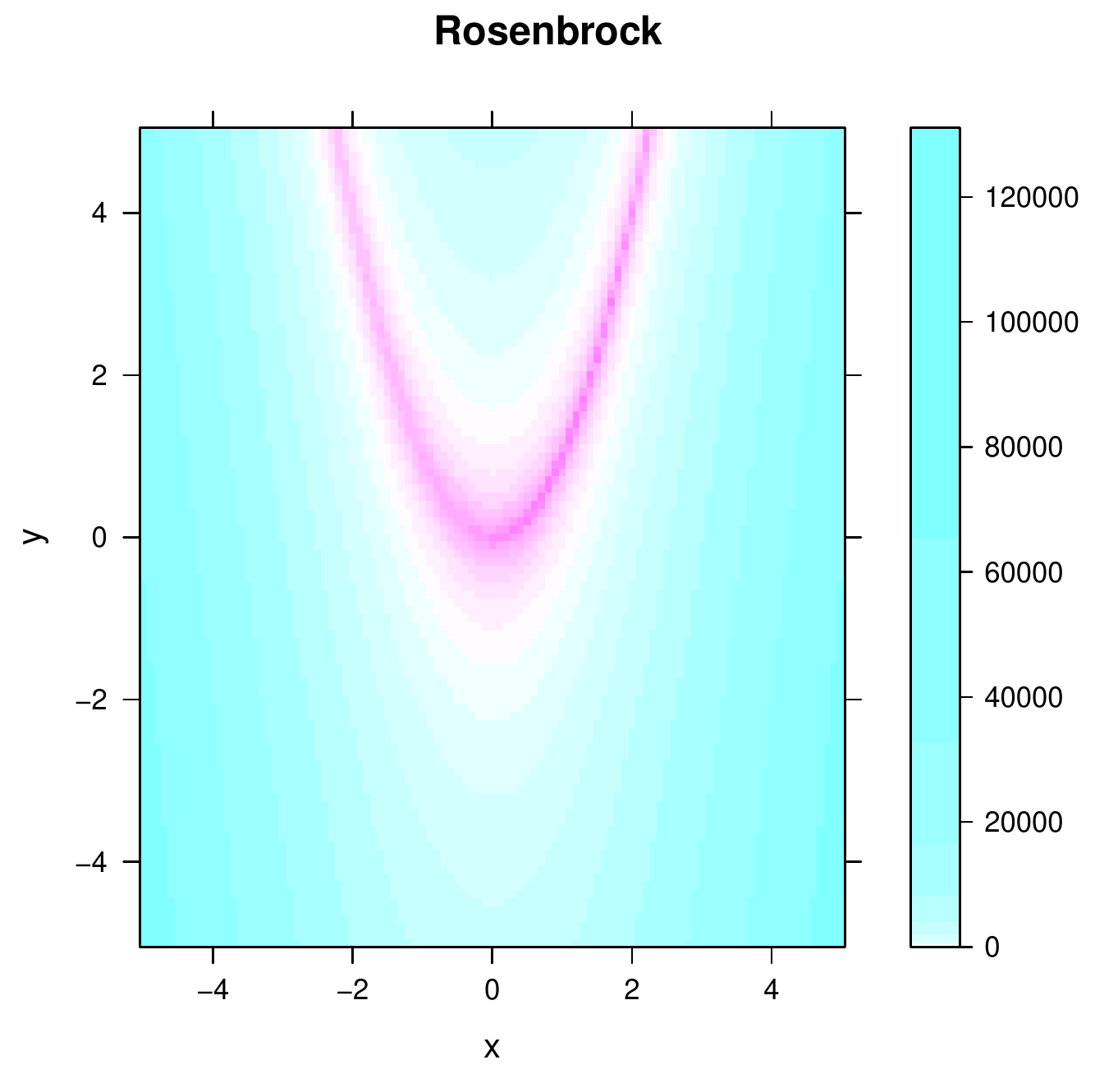} &
\includegraphics[trim={0 0 0 1cm},clip,width=1.6in]{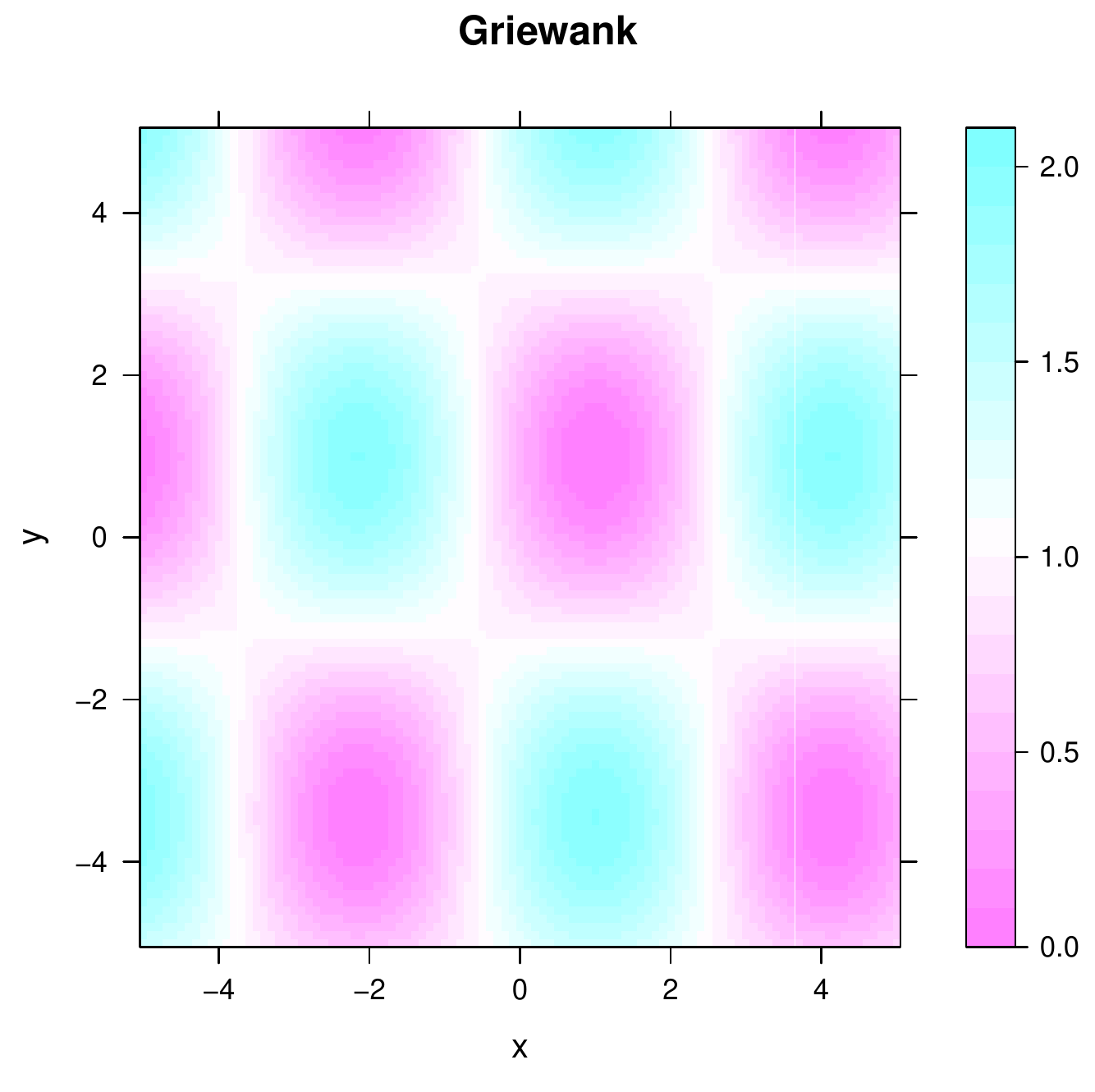} \\
Rosenbrock & Griewank
\end{tabular}
\caption{Visualization of 2-dimensional Rosenbrock and Griewank functions.}
\label{fig:visualization} 
\end{figure}

The multi-dimensional Rosenbrock function has the formula:
\begin{align}
    g(\bldx) = \sum_{i=1}^{D-1} \left(100\left(x_{i+1} - x_i^2\right)^2 + \left(1-x_i\right)^2\right).
\end{align}
The multi-dimensional Griewank function has the formula:
\begin{align}
    g(\bldx) = 1 + \frac{1}{4000}\sum_{i=1}^D\left(x_i-1\right)^2 -\prod_{i=1}^D cos\left(\frac{x_i-1}{\sqrt{i}}\right).
\end{align}
See Figure~\ref{fig:visualization} for a visulization of 2-dimensional Rosenbrock and Griewank functions. For both functions, the true global minimizer is at $\bldx^\ast = \mathbf{1}$. 

For each function, we randomly generated 50 training sets for each of 60 different experimental set-ups: $D \in \{4,8,12,16\}$ inputs $\times$ $N \in\{100,1000,10000\}$ training examples $\times$ $\sigma \in \{0.25,0.5,1.0,2.0,4.0\}$ noise levels where the training label is $\bldy=g(\bldx) + \mathbf{\epsilon}$ for $ \bldx\sim Unif(-2, 2)^D, \epsilon \sim \mathcal{N}(0, \sigma g(\bldx))$. For the conditional global optimization problem, we aim to find $\bldx^\ast = \arg\min_\bldx g(\bldx, \bldz=\bldzero)$, where $\bldx$ is the first $3D/4$ inputs and $\bldz$ is the last $D/4$ inputs. 
Once FICNN/PICNN fit their convex/conditionally-convex functions, their minimizers are found using ADAM with learning rate .001 and 10k steps with projections onto the input domain $[-2, 2]^D$. Details on hyperparameter validation for all methods are in the Appendix. 

We found the global maximizer of each response surface as in experiment in Section~\ref{sec:gonhparam}. That is, for GON and CGON, we found the global maximizer of the response surface by inverting the PLFs.  For FICNN and PICNN, we used ADAM to find their global maximizers.  For DNN and GPR, we first generated a finite random set $\mathcal{X_{\textrm{candidates}}}$ of 100,000 inputs across the domain of and set $\hat{\bldx}=\arg\max_{\bldx\in\mathcal{X_{\textrm{candidates}}}}h(\bldx)$.

Table~\ref{tab:rosenbrock} shows that GON is consistently the best method for all twelve different simulation set-ups.  CGON is also consistently best for Rosenbrock. For the globally convex Griewank, CGON is the best or tied for the best in 6 slices, whereas PICNN, DNN and GPR are the best or tied for the best in 0, 5 and 3 slices, respectively.  GON and CGON performed especially well in the more challenging cases of large $D$ and high noise $\sigma$ and few training samples $N$. 

Note that the performance of sample best deteriorates with more training samples, as there is more risk it will overfit a particularly noisy training sample. In fact, in general the performance of the different response surface methods did not necessarily get better with more training samples $N$, which we suspect is due to the fact that as $N$ increases, there is a greater chance of more very noisy samples that confuses the response surface placement of its maximizer. 

\begin{sidewaystable*}[!t]
\caption{Rosenbrock and Griewank simulation results with 95\% confidence intervals. We report $g(\hat{x})$ for each predicted minimizer $\hat{x}$, averaged over the slice that fit each row's description (e.g. all 750 runs = 50 random seeds $\times3N\times5\sigma$ where $D=4$). Lower is better for all metrics. Bold means the method is statistically significantly the best or tied for best at 95\% level. In the global optimization problem, for DNN and GPR, we take the predicted minimizer as the smallest prediction over 100,000 random sampled points from the domain.  In the conditional optimization problem, for DNN and GPR, the 100k sampled inputs are restricted to have the conditional inputs $\bldzero$. For sample best, we use $x_{i^*}$ where $i^* = \arg \min_{i =1 \ldots N} y_i$. Hyperperameters are chosen based on $g(\hat x)$ of independent runs. }
\label{tab:rosenbrock}
\centering\small
\begin{tabular}{lrrrrrrrrr}
\toprule 
& \multicolumn{5}{c}{Global Optimization} &  \multicolumn{4}{c}{Conditional Global Optimization} \\
\cmidrule(lr){2-6} \cmidrule(lr){7-10}
Rosenbrock &  GON & FICNN & DNN & GPR & Sample Best & CGON & PICNN & DNN & GPR \\
\midrule
$D=4$	&\textbf{213} $\pm$ \textit{24} & 833 $\pm$ \textit{92} & 2259 $\pm$ \textit{151} & 2310 $\pm$ \textit{186} & 3271 $\pm$ \textit{156} & \textbf{903} $\pm$ \textit{68} & 1473 $\pm$ \textit{132} & 1769 $\pm$ \textit{136} & 1558 $\pm$ \textit{157} \\
$D=8$	&\textbf{492} $\pm$ \textit{37} & 2370 $\pm$ \textit{188} & 5019 $\pm$ \textit{209} & 4791 $\pm$ \textit{241} & 5463 $\pm$ \textit{237} & \textbf{1340} $\pm$ \textit{76} & 5799 $\pm$ \textit{283} & 4334 $\pm$ \textit{201} & 3822 $\pm$ \textit{226} \\ 
$D=12$  &\textbf{734} $\pm$ \textit{47} & 3575 $\pm$ \textit{278} & 7407 $\pm$ \textit{220} & 7022 $\pm$ \textit{257} & 7559 $\pm$ \textit{273} & \textbf{1762} $\pm$ \textit{86} & 10650 $\pm$ \textit{401} & 6475 $\pm$ \textit{213} & 5430 $\pm$ \textit{256} \\
$D=16$  &\textbf{1004} $\pm$ \textit{21} & 5750 $\pm$ \textit{128} & 9466 $\pm$ \textit{91} & 9133 $\pm$ \textit{107} & 9571 $\pm$ \textit{113} & \textbf{2164} $\pm$ \textit{36} & 15682 $\pm$ \textit{177} & 8085 $\pm$ \textit{82} & 7855 $\pm$ \textit{91} \\
\midrule
$\sigma=0.25$   &\textbf{282} $\pm$ \textit{8} & 818 $\pm$ \textit{34} & 4064 $\pm$ \textit{110} & 6582 $\pm$ \textit{201} & 1576 $\pm$ \textit{59} & \textbf{943} $\pm$ \textit{22} & 5478 $\pm$ \textit{178} & 3269 $\pm$ \textit{102} & 5054 $\pm$ \textit{165} \\
$\sigma=0.5$    &\textbf{419} $\pm$ \textit{13} & 1273 $\pm$ \textit{52} & 5183 $\pm$ \textit{116} & 3830 $\pm$ \textit{117} & 6102 $\pm$ \textit{97} & \textbf{1093} $\pm$ \textit{23} & 6825 $\pm$ \textit{193} & 4260 $\pm$ \textit{101} & 3763 $\pm$ \textit{126} \\
$\sigma=1.0$    &\textbf{557} $\pm$ \textit{16} & 2805 $\pm$ \textit{97} & 6216 $\pm$ \textit{113} & 5737 $\pm$ \textit{95} & 7685 $\pm$ \textit{102} & \textbf{1500} $\pm$ \textit{32} & 8622 $\pm$ \textit{205} & 5405 $\pm$ \textit{104} & 3772 $\pm$ \textit{96} \\
$\sigma=2.0$    &\textbf{797} $\pm$ \textit{22} & 4382 $\pm$ \textit{118} & 7075 $\pm$ \textit{105} & 6445 $\pm$ \textit{77} & 8385 $\pm$ \textit{108} & \textbf{1986} $\pm$ \textit{40} & 10151 $\pm$ \textit{217} & 6184 $\pm$ \textit{98} & 5143 $\pm$ \textit{76} \\
$\sigma=4.0$    &\textbf{999} $\pm$ \textit{26} & 6383 $\pm$ \textit{139} & 7651 $\pm$ \textit{105} & 6478 $\pm$ \textit{70} & 8583 $\pm$ \textit{109} & \textbf{2188} $\pm$ \textit{44} & 10929 $\pm$ \textit{217} & 6710 $\pm$ \textit{91} & 5599 $\pm$ \textit{65} \\
\midrule
$N=100$	    &\textbf{473} $\pm$ \textit{9} & 2983 $\pm$ \textit{78} & 6462 $\pm$ \textit{83} & 5820 $\pm$ \textit{90} & 5042 $\pm$ \textit{80} & \textbf{1583} $\pm$ \textit{30} & 11407 $\pm$ \textit{173} & 5736 $\pm$ \textit{74} & 5077 $\pm$ \textit{80} \\
$N=1000$    &\textbf{897} $\pm$ \textit{19} & 4281 $\pm$ \textit{104} & 6237 $\pm$ \textit{87} & 5923 $\pm$ \textit{97} & 6481 $\pm$ \textit{93} & \textbf{1665} $\pm$ \textit{29} & 8451 $\pm$ \textit{152} & 5196 $\pm$ \textit{77} & 4777 $\pm$ \textit{87} \\
$N=10000$   &\textbf{463} $\pm$ \textit{13} & 2133 $\pm$ \textit{71} & 5414 $\pm$ \textit{97} & 5700 $\pm$ \textit{103} & 7875 $\pm$ \textit{101} & \textbf{1379} $\pm$ \textit{25} & 5345 $\pm$ \textit{133} & 4564 $\pm$ \textit{91} & 4145 $\pm$ \textit{95} \\
\midrule
\midrule
Griewank &  GON & FICNN & DNN & GPR & Sample Best & CGON & PICNN & DNN & GPR \\
\midrule
$D=4$	&\textbf{0.45} $\pm$ \textit{0.007} & 0.49 $\pm$ \textit{0.009} & 0.82 $\pm$ \textit{0.012} & 0.71 $\pm$ \textit{0.014} & 1.12 $\pm$ \textit{0.015} & 0.81 $\pm$ \textit{0.007} & 0.79 $\pm$ \textit{0.010} & 0.79 $\pm$ \textit{0.012} & \textbf{0.70} $\pm$ \textit{0.013}  \\
$D=8$	&\textbf{0.51} $\pm$ \textit{0.006} & 0.83 $\pm$ \textit{0.006} & 0.91 $\pm$ \textit{0.007} & 0.99 $\pm$ \textit{0.007} & 1.03 $\pm$ \textit{0.008} & 0.93 $\pm$ \textit{0.004} & 0.96 $\pm$ \textit{0.004} & \textbf{0.91} $\pm$ \textit{0.006} & 0.94 $\pm$ \textit{0.008} \\ 
$D=12$  &\textbf{0.59} $\pm$ \textit{0.005} & 0.90 $\pm$ \textit{0.004} & 0.97 $\pm$ \textit{0.004} & 1.02 $\pm$ \textit{0.002} & 1.04 $\pm$ \textit{0.004} & 0.97 $\pm$ \textit{0.003} & 1.00 $\pm$ \textit{0.002} & \textbf{0.96} $\pm$ \textit{0.004} & 1.01 $\pm$ \textit{0.003} \\
$D=16$  &\textbf{0.64} $\pm$ \textit{0.005} & 0.94 $\pm$ \textit{0.003} & 1.00 $\pm$ \textit{0.002} & 1.01 $\pm$ \textit{0.002} & 1.02 $\pm$ \textit{0.003} & \textbf{0.97} $\pm$ \textit{0.002} & 1.01 $\pm$ \textit{0.001} & 0.99 $\pm$ \textit{0.003} & 1.01 $\pm$ \textit{0.002} \\
\midrule
$\sigma=0.25$   &\textbf{0.53} $\pm$ \textit{0.006} & 0.65 $\pm$ \textit{0.008} & 0.73 $\pm$ \textit{0.010} & 0.74 $\pm$ \textit{0.012} & 0.68 $\pm$ \textit{0.012} & 0.87 $\pm$ \textit{0.005} & 0.86 $\pm$ \textit{0.008} & \textbf{0.72} $\pm$ \textit{0.010} & 0.76 $\pm$ \textit{0.011} \\
$\sigma=0.5$    &\textbf{0.52} $\pm$ \textit{0.006} & 0.71 $\pm$ \textit{0.008} & 0.86 $\pm$ \textit{0.009} & 0.80 $\pm$ \textit{0.011} & 1.07 $\pm$ \textit{0.006} & 0.91 $\pm$ \textit{0.005} & 0.91 $\pm$ \textit{0.007} & 0.86 $\pm$ \textit{0.008} & \textbf{0.78} $\pm$ \textit{0.010} \\
$\sigma=1.0$    &\textbf{0.52} $\pm$ \textit{0.006} & 0.79 $\pm$ \textit{0.008} & 0.95 $\pm$ \textit{0.007} & 0.95 $\pm$ \textit{0.009} & 1.13 $\pm$ \textit{0.007} & \textbf{0.93} $\pm$ \textit{0.005} & 0.95 $\pm$ \textit{0.006} & \textbf{0.93} $\pm$ \textit{0.007} & \textbf{0.92} $\pm$ \textit{0.009} \\
$\sigma=2.0$    &\textbf{0.58} $\pm$ \textit{0.008} & 0.86 $\pm$ \textit{0.008} & 1.03 $\pm$ \textit{0.005} & 1.06 $\pm$ \textit{0.006} & 1.17 $\pm$ \textit{0.008} & \textbf{0.94} $\pm$ \textit{0.005} & 0.98 $\pm$ \textit{0.006} & 1.00 $\pm$ \textit{0.006} & 1.03 $\pm$ \textit{0.007} \\
$\sigma=4.0$    &\textbf{0.60} $\pm$ \textit{0.008} & 0.93 $\pm$ \textit{0.007} & 1.05 $\pm$ \textit{0.005} & 1.10 $\pm$ \textit{0.006} & 1.19 $\pm$ \textit{0.008} & \textbf{0.95} $\pm$ \textit{0.006} & 1.00 $\pm$ \textit{0.005} & 1.05 $\pm$ \textit{0.006} & 1.09 $\pm$ \textit{0.006} \\
\midrule
$N=100$	    &\textbf{0.59} $\pm$ \textit{0.006} & 0.91 $\pm$ \textit{0.006} & 0.99 $\pm$ \textit{0.004} & 1.00 $\pm$ \textit{0.005} & 0.98 $\pm$ \textit{0.006} & \textbf{0.95} $\pm$ \textit{0.005} & 0.98 $\pm$ \textit{0.004} & 0.98 $\pm$ \textit{0.004} & 0.99 $\pm$ \textit{0.005} \\
$N=1000$    &\textbf{0.52} $\pm$ \textit{0.005} & 0.79 $\pm$ \textit{0.007} & 0.98 $\pm$ \textit{0.005} & 0.94 $\pm$ \textit{0.008} & 1.04 $\pm$ \textit{0.008} & \textbf{0.91} $\pm$ \textit{0.004} & 0.97 $\pm$ \textit{0.005} & 0.97 $\pm$ \textit{0.006} & 0.92 $\pm$ \textit{0.007} \\
$N=10000$   &\textbf{0.53} $\pm$ \textit{0.004} & 0.66 $\pm$ \textit{0.006} & 0.81 $\pm$ \textit{0.008} & 0.85 $\pm$ \textit{0.010} & 1.13 $\pm$ \textit{0.008} & 0.90 $\pm$ \textit{0.004} & 0.87 $\pm$ \textit{0.006} & \textbf{0.79} $\pm$ \textit{0.008} & 0.84 $\pm$ \textit{0.009} \\
\bottomrule
\end{tabular}
\end{sidewaystable*}

The multi-dimensional Rosenbrock function has formula:
\begin{align}
    g(\bldx) = \sum_{i=1}^{D-1} \left(100\left(x_{i+1} - x_i^2\right)^2 + \left(1-x_i\right)^2\right).
\end{align}
The multi-dimensional Griewank function has formula:
\begin{align}
    g(\bldx) = 1 + \frac{1}{4000}\sum_{i=1}^D\left(x_i-1\right)^2 -\prod_{i=1}^D cos\left(\frac{x_i-1}{\sqrt{i}}\right).
\end{align}
For both functions, the true global minimizer is at $\bldx^\ast = (1.0, 1.0, \dots, 1.0)$.

For both GON and CGON, we first use D PLFs with $K$ keypoints to calibrate the D inputs for optimization. The unimodal function consists of an enesemble of $D$ unimodal lattices, each fuses 3 inputs with $V$ keypoints.
For CGON, we let $r:\mathbb{R}^M\to\calS_D$ be $r(\bldz)[j] = \sum_{i=1}^M PLF_i^j(\bldz[i]), j = 1,\dots,D$, where $\bldz[i]$ and $r(z)[i]$ denote the $i$-th entry of $\bldz$ and $r(\bldz)$.

For FICNN and PICNN, we use the formulations in (2) (Figure 1) and (3) (Figure 2), respectively, in Amos, et al. \cite{Zico:2017}. All the hidden layers are constructed to have the same hidden dimensions, whenever possible. For DNN, we use fully connected hidden layers with a constant number of hidden nodes across layers. The number of hidden layers and the number of hidden nodes are treated as hyperparameters.

For GPR, we use RBF kernel with $\sigma = 1$, which is the default in the sklearn package. The White Kernel $\alpha$ is treated as a hyperparameter. 

For each of the Rosenbrock and Griewank functions, we used grid search to choose hyperparameters for each method that minimize the average $g(\hat{\bldx})$ over the $600$ runs for each function ($5 \sigma\times4D\times3N\times10$ repetitions with random seeds), where $g$ denotes the ground truth function. After choosing hyperparameters, we reran the simulation with 50 repetitions, and report the average-50 result in Table~\ref{sec:gonsims}. The hyperprameters of each method are summarized in Table~\ref{tab:hparamsims}. 

\begin{table*}[!t]
\caption{Simulation: hyperparameters.}
\label{tab:hparamsims}
\centering
\begin{tabular}{lrrrrrrrr}
\toprule 
& \multicolumn{4}{c}{Global Optimization} &  \multicolumn{4}{c}{Conditional Global Optimization} \\
\cmidrule(lr){2-5} \cmidrule(lr){6-9}
Rosenbrock &  GON & FICNN & DNN & GPR &  CGON & PICNN & DNN & GPR \\
\midrule
PLF kps per input $K$ & 10 & - & - & - & 10 & - & - & - \\
Lattice kps per input $V$ & 3 & - & - & - & 3 & - & - & - \\
Num hidden layers & - & 2 & 2 & - & - & 2 & 2 & - \\
Num hidden nodes & - & 16 & 32 & - & - & 16 & 16 & - \\
$\alpha$ in GPR  & - & - & - & 1.0 & - & - & - & 1.0 \\
\midrule
\midrule
Griewank &  GON & FICNN & DNN & GPR &  CGON & PICNN & DNN & GPR \\
\midrule
PLF kps per input $K$ & 10 & - & - & - & 10 & - & - & - \\
Lattice kps per input $V$ & 3 & - & - & - & 3 & - & - & - \\
Num hidden layers & - & 4 & 2 & - & - & 2 & 2 & - \\
Num hidden nodes & - & 32 & 16 & - & - & 16 & 16 & - \\
$\alpha$ in GPR  & - & - & - & 1.0 & - & - & - & 1.0 \\
\bottomrule
\end{tabular}
\end{table*}

\section{Open Questions}

We defined GONs (and CGONs) by the shape constraints they must obey: a composition of invertible layers and unimodal layers. We showed how to construct such models using the piece-wise linear functions and lattice layers of DLNs, which are arbitrarily flexible models that are particularly amenable to shape constraints \citep{Gupta:2020,SetConstraintsICML2019}, but other functions could be used for the invertible layers \cite{Behrmann:2019}, and one could use convex networks for the needed unimodal layers (at the cost of some flexibility) \citep{Zico:2017}. 

Another open question is the choice of loss function when training GONs or other flexible response surfaces. In our experiments, all models were fit using standard mean-squared error. Since the goal of fitting the GON is to predict the maximizer only, it seems intuitive that one should worry more about fitting the examples closer to the (unknown) maximizer. We experimented with loss functions that up-weighted training examples with bigger label values, but, perhaps due to the flexibility of the GONs, did not find they helped much, and eschewed their extra complexity and hyperparameters.  

We focused here on the setting where one makes only one prediction. However  GONs could also be used as a response surface function within a global optimization algorithm that is able to make a series of guesses. In such a context it might make sense to evolve the neighborhood fit by the GON, or fit many GONs in parallel to different evolving neighborhoods for a multi-agent search like particle-swarm optimization \cite{Kennedy:1995, Shi:1998}. 

Lastly, this work is part of a recent wave of research into shape constraints showing that shape constraints can provide sensible regularization while not hurting useful expressability of AI models (e.g. \cite{Pya:2015,Chen:2016,GuptaEtAl:2016,Cannon:2018,Chetverikov:2018,SetConstraintsICML2019,Louppe:2019,gasthaus2019,WangGupta:2020}).   We hope this work will inspire other useful shape constraint regularization strategies for AI.

\end{document}